%% file: main.tex
\documentclass[letterpaper,journal]{IEEEtran}

\usepackage[T1]{fontenc}
\usepackage[utf8]{inputenc}
\usepackage{microtype}

\usepackage{amsmath,amsfonts,amssymb}

\usepackage{graphicx}
\usepackage{booktabs}
\usepackage{array}
\usepackage{multirow}
\usepackage{adjustbox}
\usepackage{tabularx}
\usepackage[table]{xcolor}
\usepackage[caption=false,font=normalsize,labelfont=sf,textfont=sf]{subfig}
\usepackage{textcomp}
\usepackage{stfloats}
\usepackage{placeins}
\usepackage{url}
\usepackage{cite}
\usepackage{algorithm}
\usepackage{algpseudocode}
\algrenewcommand\algorithmiccomment[1]{\hfill \(\triangleright\) \textcolor{blue}{#1}}

\usepackage[most]{tcolorbox}
\usepackage{listings}
\usepackage[hidelinks]{hyperref}

\definecolor{gptbar}{RGB}{232,244,255}
\definecolor{qwenbar}{RGB}{242,235,255}
\definecolor{llamabar}{RGB}{255,251,214}

\definecolor{caseblue}{RGB}{232,242,252}
\definecolor{casegray}{RGB}{248,248,248}
\definecolor{caseborder}{RGB}{130,130,130}
\definecolor{methodrow}{RGB}{228,240,250}
\definecolor{tablegroup}{RGB}{242,242,242}

\newcolumntype{Y}{>{\raggedright\arraybackslash}X}

\newcommand{\method}{\textsc{QGMem}}
\newcommand{\citep}[1]{\cite{#1}}
\newcommand{\citet}[1]{\cite{#1}}

\makeatletter
\newcommand{\compactfigurecaption}[2]{%
  \begingroup
  \def\@captype{figure}%
  \refstepcounter{figure}%
  \def\@currentlabelname{#2}%
  \label{#1}%
  \addcontentsline{lof}{figure}{\protect\numberline{\thefigure}#2}%
  \@makecaption{\fnum@figure}{#2}%
  \endgroup
}
\newcommand{\compacttablecaption}[2]{%
  \begingroup
  \def\@captype{table}%
  \refstepcounter{table}%
  \def\@currentlabelname{#2}%
  \label{#1}%
  \addcontentsline{lot}{table}{\protect\numberline{\thetable}#2}%
  \@makecaption{\fnum@table}{#2}%
  \endgroup
}
\newcommand{\fixedsection}{\@startsection{section}{1}{\z@}{3.0ex}%
{0.7ex plus 1ex minus 0ex}{\normalfont\normalsize\centering\scshape}}
\newcommand{\fixedsubsection}{\@startsection{subsection}{2}{\z@}{3.5ex}%
{0.4ex}{\normalfont\normalsize\itshape}}
\newcommand{\tightsubsection}{\@startsection{subsection}{2}{\z@}{0.5ex}%
{0.2ex}{\normalfont\normalsize\itshape}}
\newcommand{\fixedsubsubsection}{\@startsection{subsubsection}{3}{\parindent}{0pt}%
{0pt}{\normalfont\normalsize\itshape}}
\makeatother

\lstdefinestyle{promptstyle}{
  basicstyle=\ttfamily\scriptsize,
  breaklines=true,
  breakatwhitespace=true,
  columns=fullflexible,
  keepspaces=true,
  xleftmargin=0pt,
  framexleftmargin=0pt,
  aboveskip=0pt,
  belowskip=0pt,
  resetmargins=true
}

\newtcolorbox{casebox}[1][]{
  enhanced,
  colback=white,
  colframe=caseborder,
  boxrule=0.45pt,
  arc=1.2mm,
  left=1.2mm,
  right=1.2mm,
  top=0.8mm,
  bottom=0.8mm,
  #1
}

\newtcbox{\keybox}{
  enhanced,
  on line,
  colback=caseblue,
  colframe=caseborder,
  boxrule=0.3pt,
  arc=1mm,
  width=5mm,
  height=2.6mm,
  left=0pt,
  right=0pt,
  top=0pt,
  bottom=0pt
}

\title{Event-Centric Memory with Query-Aware Graph Augmentation for Long-Term Conversational Agents}

\author{Yichen~Liu,
        Chunfeng~Yuan,
        Haowei~Liu,
        Wenjuan~Li,
        Zefeng~Lin,
        Bing~Li,
        Xu~Chen,
        and~Weiming~Hu%
\thanks{\makebox[0.95\linewidth][c]{\rule{0.55\linewidth}{0.4pt}}\par\vspace{0.35em}\itshape\hangindent=1.5em\hangafter=1 $\bullet$\enspace Y. Liu, C. Yuan, W. Li, Z. Lin, B. Li, and W. Hu are with the State Key Laboratory of Multimodal Artificial Intelligence Systems, Institute of Automation, Chinese Academy of Sciences, Beijing 100190, China.}%
\thanks{\itshape\hangindent=1.5em\hangafter=1 $\bullet$\enspace Y. Liu, C. Yuan, B. Li, and W. Hu are also with the School of Artificial Intelligence, University of Chinese Academy of Sciences, Beijing 100049, China.}%
\thanks{\itshape\hangindent=1.5em\hangafter=1 $\bullet$\enspace C. Yuan, W. Li, Z. Lin, B. Li, and W. Hu are also with the Beijing Key Laboratory of Super Intelligent Security of Multi-Modal Information, Beijing, China.}%
\thanks{\itshape\hangindent=1.5em\hangafter=1 $\bullet$\enspace H. Liu is with Alibaba Group, China.}%
\thanks{\itshape\hangindent=1.5em\hangafter=1 $\bullet$\enspace B. Li is also with PeopleAI Inc., Beijing 100190, China.}%
\thanks{\itshape\hangindent=1.5em\hangafter=1 $\bullet$\enspace X. Chen is with the Gaoling School of Artificial Intelligence, Renmin University of China, Beijing 100872, China.}%
\thanks{\itshape\hangindent=1.5em\hangafter=1 $\bullet$\enspace W. Hu is also with the School of Information Science and Technology, ShanghaiTech University, Shanghai 201210, China.}%
\thanks{\itshape\hangindent=1.5em\hangafter=1 $\bullet$\enspace Corresponding author: Chunfeng Yuan (e-mail: cfyuan@nlpr.ia.ac.cn).}%
}

\begin{document}
\maketitle

\begin{abstract}
For persistent and personalized conversational agents, memory systems can enable them to remember, update, and reason over long histories by storing past interactions and retrieving relevant information. Existing memory systems typically follow two paradigms: flat-structured memory and graph-based memory. The former is lightweight but leaves event relations and state updates implicit, while the latter explicitly models memory structure but incurs additional construction cost and introduces irrelevant relations over long histories. To address these limitations, we propose \method{}, a novel memory construction and activation framework motivated by human memory, in which experience is organized into events and query-relevant events are modeled by graph as working memory. \method{} converts long dialogue histories into event-indexed atomic memory units that preserve individual experiences and consolidates related units into dynamic memory traces that retain state trajectories and current states. When a query arrives, hybrid memory retrieval gathers complementary candidate memories, and query-aware reranking activates the most relevant units as a compact working memory. To expose relational dependencies in the working memory and support conflict-aware reasoning, \method{} organizes the working memory as a local graph, which is then encoded as a graph token and provided to the LLM together with the textual working memory to improve evidence utilization during answer generation. Experiments across six benchmarks validate the framework and show consistent gains in retrieval, multi-hop evidence composition, conflict resolution, and ultra-long dialogue reasoning with compact contexts and moderate inference cost.

\end{abstract}


\section{Introduction}

Maintaining continuity across long-term interaction between humans and agents remains a pivotal milestone towards personalized and autonomous artificial intelligence. Personal assistants, companion dialogue systems, and task-oriented agents are expected to remember user preferences, interpersonal relations, schedules, and later updates across many turns or sessions. Since such information cannot always fit into the model context~\citep{liu2024lostmiddle}, an agent needs memory mechanisms that preserve past interactions, track evolving states, identify relevant memory, and organize their relations for answer generation. Therefore, recent works~\citep{wang2024survey,zhong2024memorybank,packer2023memgpt} introduce long-term memory systems that store historical interactions as persistent external memory and retrieve relevant memory units when answering user queries.

\begin{figure*}[!t]
\centering
\includegraphics[width=\textwidth]{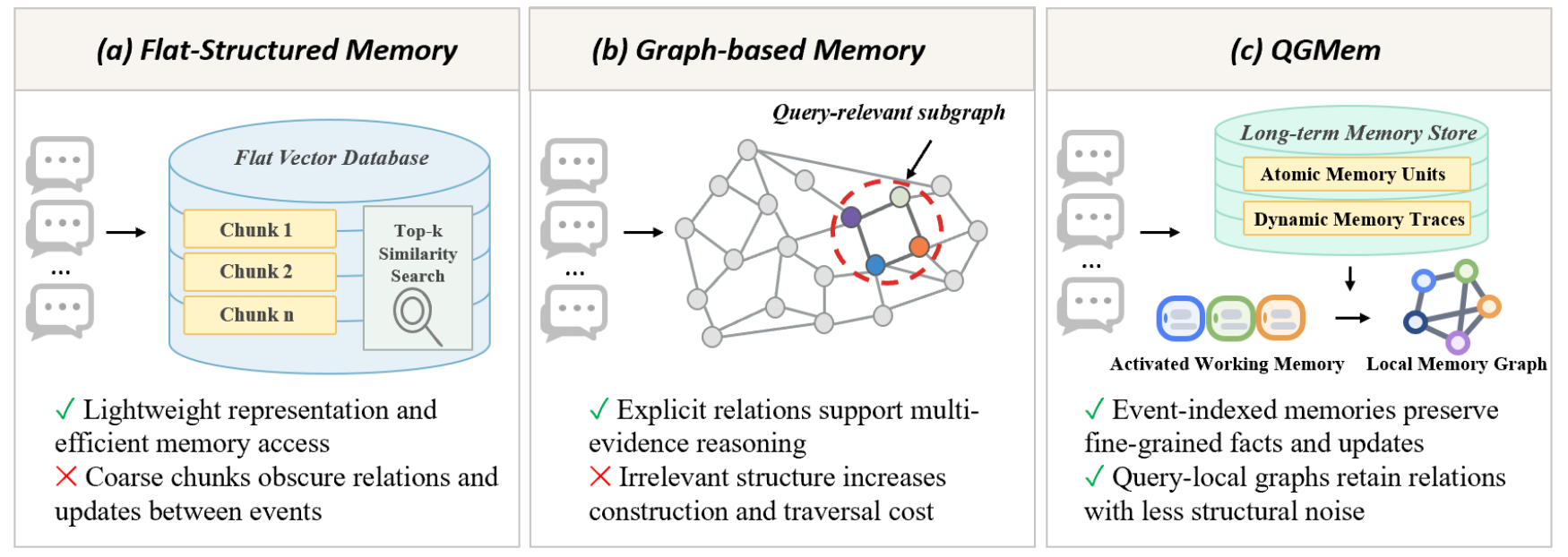}
\caption{Comparison of flat-structured memory, graph-based memory, and \method{}.
}
\label{fig:intro-compare}
\end{figure*}

Existing memory systems mainly follow two organizational paradigms, as illustrated in Figure~\ref{fig:intro-compare}(a) and (b). \textbf{Flat-structured memory} stores historical information as turns, sessions or summaries~\citep{zhong2024memorybank,packer2023memgpt,wang2024fact}. This paradigm is lightweight and easy to maintain, but relations among events and state changes are usually left implicit. \textbf{Graph-based memory} explicitly represents entities, events, and states as graph structures~\citep{wu2025sgmem,zhou2025emem,yang2026graphmemory}. Such structures support relational reasoning, but constructing and retrieving graph structures over long histories introduces additional cost and even irrelevant relations. Consequently, existing long-term conversational memory systems struggle to simultaneously achieve lightweight persistent storage and explicit relational organization.

These limitations motivate a memory architecture that combines persistent storage with query-time relational organization: long-term memory preserves events and their evolving states, whereas working memory constructs a relational structure over query-relevant events. This view is consistent with classic research in cognitive psychology on event cognition, which suggests that people represent continuous experience as a sequence of events, update the current event in working memory, and retrieve relevant past events from long-term memory~\citep{zacks2007event,radvansky2012across,radvansky2017event}. Motivated by this perspective, we propose \underline{Q}uery-aware \underline{G}raph-augmented \underline{Mem}ory (\method{}), which combines event-centric memory storage with query-time relational organization, as illustrated in Figure~\ref{fig:intro-compare}(c). \method{} first converts dialogue histories into event-indexed atomic memory units with traceable source content, each representing an independently interpretable event or state and indexed by explicit event attributes, including entities, locations, and factual time. Related units are then consolidated into dynamic memory traces that preserve state trajectories and current states, thereby making state updates explicit and distinguishing current facts from superseded ones. To improve the retrieval quality of answer-bearing memories, hybrid memory retrieval integrates lexical, semantic, and event-attribute relevance using reciprocal rank fusion (RRF)~\citep{cormack2009rrf}. Query-aware reranking removes irrelevant and superseded candidates, yielding a compact working memory concentrated on the current query. To model relational dependencies within the activated working memory, \method{} constructs a local memory graph from the facts of the selected memory units. Then, a relation-aware graph encoder is employed to encode the local memory graph, and the resulting graph-level representation is projected into the LLM embedding space. The graph token is supplied to the LLM together with the textual working memory, enabling multi-hop evidence composition based on both memory content and relational information.

\begin{figure}[!t]
\centering
\includegraphics[width=0.92\columnwidth]{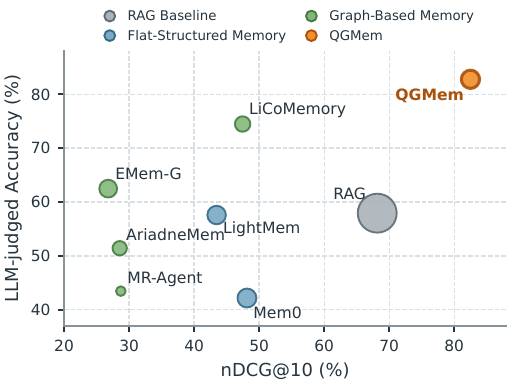}
\caption{Comparison of retrieval quality, LLM-judged answer score, and retrieved-context size on LoCoMo with Qwen3-8B. Colors distinguish the retrieval-augmented baseline, flat-structured memory, graph-based memory, and \method{}. The horizontal axis denotes normalized discounted cumulative gain at rank 10 (nDCG@10), the vertical axis denotes the LLM-judged score, and bubble area denotes the average number of retrieved-context tokens. Higher axis values and smaller bubble areas indicate better overall performance.}
\label{fig:locomo-qwen-bubble}
\end{figure}

The resulting framework is designed to reconcile three objectives in long-term conversational memory: preserving verifiable historical evidence, exposing query-relevant relations, and maintaining a compact generation context. Figure~\ref{fig:locomo-qwen-bubble} provides an empirical view of this balance by comparing \method{} with retrieval-augmented generation (RAG) and representative flat-structured and graph-based memory systems in terms of retrieval quality, answer quality, and context efficiency. Flat-structured systems maintain compact contexts but exhibit lower retrieval or answer quality, whereas graph-based systems do not simultaneously attain high retrieval and answer quality. In contrast, \method{} achieves an nDCG@10 of 82.50\% and an LLM-judged score of 82.77\% with 598.19 retrieved-context tokens. These results show that \method{} improves the prioritization and utilization of answer-supporting memories while maintaining a compact context for generation.

We conduct main experiments on six long-term memory benchmarks. LoCoMo~\citep{maharana2024locomo} and LongMemEval-S~\citep{wu2025longmemeval} evaluate accurate retrieval and multi-hop evidence composition, while Fact-SH and Fact-MH from MemoryAgentBench~\citep{hu2025memoryagentbench} evaluate conflict resolution. BEAM-100K and BEAM-500K~\citep{tavakoli2026beam} assess scalability over ultra-long dialogue histories. We further extend the BEAM evaluation to 1M and 10M tokens to examine performance under extreme context lengths. Overall, \method{} consistently outperforms state-of-the-art systems. Further evaluations show that these gains remain consistent across two open-weight backbones, Qwen3-8B and Llama3.1-8B-Instruct, and one closed-source backbone, GPT-4o-mini~\citep{yang2025qwen3,grattafiori2024llama3,openai2024gpt4omini}. In-depth ablation experiments confirm the effectiveness of the key components in \method{}. Additional efficiency, scalability, hyperparameter sensitivity, and case analyses characterize the computational cost, robustness, and retrieval performance of \method{} from complementary perspectives.

The main contributions of this work are summarized as follows:
\begin{itemize}
    \item We introduce an event-centric memory framework for persistent conversations that integrates fine-grained event storage, evolving-state consolidation, query-aware activation, and relational evidence organization within a unified memory lifecycle.
    \item We represent conversational experience using traceable atomic memory units and dynamic memory traces, preserving both historical states and currently valid information as the dialogue evolves.
    \item We organize query-relevant memories into a compact working memory and encode their local relational topology jointly with source-grounded textual evidence, supporting evidence composition without maintaining a graph over the complete dialogue history.
    \item We evaluate the framework across six benchmarks and multiple LLM backbones, examining answer quality, retrieval effectiveness, conflict-sensitive reasoning, long-context scalability, and computational cost.
\end{itemize}

\section{Related Work}
\label{sec:related-work}

\subsection{Persistent Conversational Memory Representation}
A central design choice in conversational memory is how interaction histories are represented and organized for persistent storage~\citep{wang2024survey,zhang2025agentmemorysurvey,hu2026memorysurvey,yang2026graphmemory}. Flat-structured memory retains dialogue turns or transforms them into chunks, summaries, and facts. MemGPT manages prior interactions through external memory~\citep{packer2023memgpt}, while LongMem maintains a retrievable long-term memory cache~\citep{wang2023longmem}. MemoryBank and FACT organize summarized or extracted information~\citep{zhong2024memorybank,wang2024fact}, whereas Mem0 and LightMem consolidate salient records to improve compactness and access efficiency~\citep{chhikara2025mem0,fang2026lightmem}. These representations are easy to maintain and can be supplied directly to an LLM, but event relations and state changes remain implicit.

Graph-based conversational memory, by contrast, materializes explicit associations at different granularities. A-MEM organizes memory notes into an evolving network through dynamically generated links and updates~\citep{xu2025amem}. Associa constructs an event-centric memory graph and combines subgraph extraction with iterative deliberative recall~\citep{zhang2025associa}. Zep maintains a temporal knowledge graph that records the validity and history of relations~\citep{rasmussen2025zep}; SG-Mem connects sentences within chunked dialogue units~\citep{wu2025sgmem}; and EMem-G organizes sessions, event-like discourse units, and their arguments in a heterogeneous graph~\citep{zhou2025emem}. These approaches differ in graph granularity and edge semantics, while sharing the use of an explicit graph as a persistent memory index.

Recent systems further enrich persistent graphs with hierarchical, temporal, and provenance-aware organization. LiCoMemory and GAM connect detailed events with higher-level semantic structures~\citep{huang2026licomemory,wu2026gam}, while HyperMem uses hyperedges to capture higher-order associations among topics, episodes, and facts~\citep{yue2026hypermem}. APEX-MEM adopts an append-only temporal property graph to preserve evolving facts~\citep{banerjee2026apexmem}, and MemORAI couples a provenance-enriched multi-relational graph with query-adaptive edge weighting~\citep{van2026memorai}. Complementary structured-memory approaches such as StructMem preserve event-level bindings and periodically consolidate cross-event connections without imposing a persistent entity--relation graph~\citep{xu2026structmem}. A controlled analysis of long-term dialogue memory systems further shows that graph effectiveness depends jointly on memory-unit design, graph maintenance, seed activation, and graph expansion, making the surrounding retrieval pipeline important for interpreting gains from graph structure~\citep{hu2026memorygraphs}.

Persistent graphs expose relational structure throughout the memory lifecycle, but relation extraction, alignment, and maintenance operate over an expanding history. \method{} places graph materialization after query-aware activation. Its persistent substrate consists of source-grounded atomic units and lightweight dynamic traces; a relational graph is instantiated only over the activated working memory. Dynamic traces preserve longitudinal state evolution, while the query-local graph organizes dependencies among the selected facts. This separation limits the scope of graph construction without discarding historical evidence or its provenance.

\subsection{Query-Time Memory Access}
Effective use of persistent conversational memory depends on retrieving information that is relevant and sufficient for the current query without introducing extraneous context. Flat-structured memory systems generally rank independent records by lexical or semantic relevance. Lexical retrieval, commonly implemented with BM25~\citep{robertson2009bm25}, preserves exact names, dates, and other discriminative expressions, but misses records expressed in substantially different terms. Dense retrieval captures semantic correspondence beyond exact terms~\citep{reimers2019sbert,karpukhin2020dpr,izacard2022contriever} and is used by systems such as LongMem and Mem0~\citep{wang2023longmem,chhikara2025mem0}. However, descriptions of similar events may receive high semantic relevance despite referring to different entities, times, or states.

Retrieval over a persistent graph commonly begins with semantically matched seed nodes and then expands through stored edges~\citep{hu2026memorygraphs}. Associa extracts evidence-rich subgraphs through prize-collecting Steiner tree optimization~\citep{zhang2025associa}; AriadneMem discovers bridge paths between distributed evidence~\citep{zhu2026ariadnemem}; HyperMem performs coarse-to-fine retrieval over its topic, episode, and fact hierarchy~\citep{yue2026hypermem}; and GAM follows cross-level links from relevant topics to archived event graphs~\citep{wu2026gam}. MemORAI applies query-conditioned edge weighting during graph propagation~\citep{van2026memorai}, while APEX-MEM uses a retrieval agent to resolve conflicting and evolving facts~\citep{banerjee2026apexmem}. These strategies exploit stored topology to broaden evidence coverage, while their retrieval behavior remains coupled to the construction and maintenance of the persistent graph.

To address these limitations, \method{} applies hybrid retrieval over event-indexed atomic memory units to improve the coverage of answer-supporting information while reducing the retrieval of events whose attributes do not match the query. Query-aware reranking then filters irrelevant and superseded candidates to form a compact working memory. A local graph constructed over this activated memory preserves explicit relational organization for evidence composition while avoiding the maintenance cost and query-irrelevant relations associated with a persistent global graph.

\subsection{Memory-Augmented Language Generation}
Beyond selecting relevant records, memory-augmented generation determines how retrieved information is incorporated into the LLM. Most systems provide retrieved memories to the LLM as text: flat records are appended directly to the prompt, while graph-based memory is serialized as triples, paths, subgraphs, or summaries~\citep{edge2025graphrag,gutierrez2024hipporag,rasmussen2025zep,zhu2026ariadnemem}. Although broadly compatible with language models, serialization linearizes graph topology, requiring the LLM to reconstruct structural dependencies from tokens and increasing context consumption.

Recent studies on integrating graph structures with LLMs have shown that learned graph representations can preserve topology and provide relational signals directly during generation, without requiring the LLM to reconstruct these dependencies solely from serialized text~\citep{tang2024graphgpt,chen2024llaga}. G-Retriever combines an encoded subgraph with its textual representation for graph question answering~\citep{he2024gretriever}, while RAS and Relink construct query-specific evidence graphs for knowledge-intensive generation and GraphRAG~\citep{jiang2026ras,huang2026relink}. \method{} builds on the general principle of query-conditioned graph augmentation but addresses persistent conversational memory, where facts may be introduced, revised, supplemented, or superseded over time. It therefore couples query-local relational organization with event-indexed atomic storage and dynamic memory traces, allowing generation to use selected evidence together with its retained state history.

\fixedsection{Method}

\begin{figure*}[t]
\centering
\includegraphics[width=\textwidth]{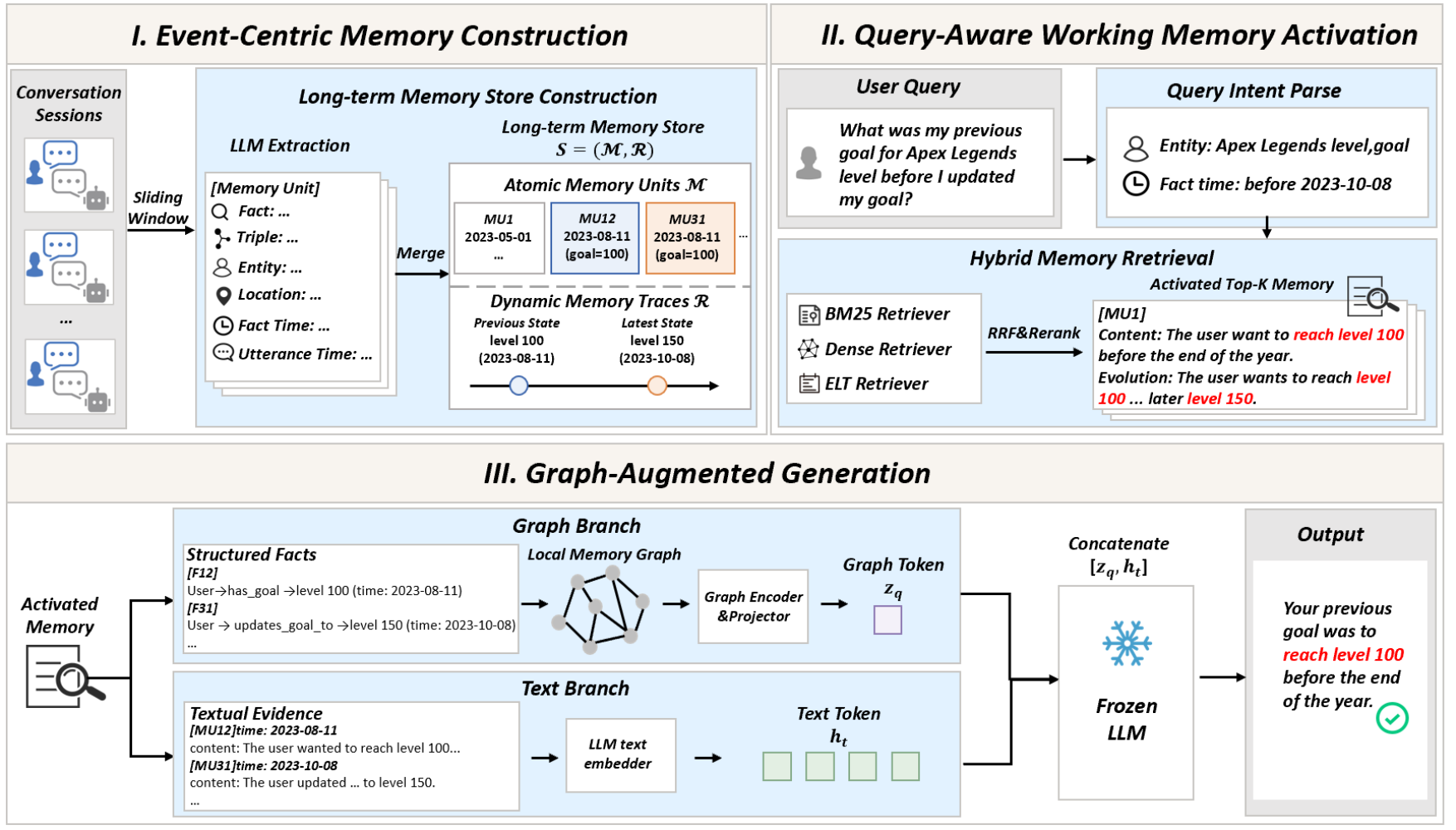}
\caption{Overall framework of \method{}.}
\label{fig:overview}
\end{figure*}

The overall framework of \method{} is illustrated in Figure~\ref{fig:overview} and comprises three modules. First, \textbf{event-centric memory construction} converts the dialogue history into event-indexed atomic memory units and consolidates related units into dynamic memory traces, preserving verifiable evidence and evolving states. Second, \textbf{query-aware working memory activation} retrieves and reranks memories for the query \(q\), forming a compact working memory that concentrates complete, current, and answer-relevant information. Finally, \textbf{graph-augmented generation} organizes the activated memories into a local graph and jointly provides its graph token and textual working memory to the LLM, enabling answer generation from both memory content and structural relations. In the following, we will introduce these three modules in detail, and analyze the generalization and efficiency of our proposed framework.

\fixedsubsection{Event-Centric Memory Construction}

\method{} represents a dialogue history as an event-indexed atomic memory set \(\mathcal{M}\) and a dynamic trace set \(\mathcal{R}\). Each atomic unit represents a fine-grained event or state and retains its source dialogue, structured fact, and event and temporal attributes for traceable retrieval. Dynamic traces connect related units into state trajectories that distinguish current information from superseded records. Together, they form the long-term memory store \(\mathcal{S}=(\mathcal{M},\mathcal{R})\), supporting precise memory access and conflict-aware answering over evolving histories.
  
\subsubsection{Atomic Memory Unit Construction}
To construct atomic memories, \method{} first scans each dialogue session with \(N_W\) overlapping sliding windows \(\{W_i\}_{i=1}^{N_W}\). Each window \(W_i\) provides a local dialogue context that contains newly processed turns together with limited preceding context for resolving pronouns and local references. Given \(W_i\), \method{} decomposes the dialogue into independently retrievable memory units with traceable source content, each representing a single interpretable event or state. When one extracted candidate contains multiple independently retrievable relations, they are materialized as separate atomic units so that each finalized unit is anchored by one complete structured fact. Let \(K_i\) be the number of units extracted from \(W_i\), and let \(m_{i,j}\) denote its \(j\)-th unit. The complete memory set is \(\mathcal{M}=\{m_{i,j}\mid 1\leq i\leq N_W,1\leq j\leq K_i\}\). For a generic unit \(m\in\mathcal{M}\), \(c_m\) denotes its context-independent summary, \(d_m\) the corresponding source dialogue span, \(\mathcal{E}_m\) and \(\mathcal{P}_m\) its entity and location sets, and \(f_m\) its structured fact. The factual and utterance times of the unit are denoted by \(T_m^f\) and \(T_m^u\). The unit is represented as
\begin{equation}
m=(c_m,d_m,\mathcal{E}_m,\mathcal{P}_m,T_m^f,T_m^u,f_m).
\end{equation}
Together, these fields preserve source traceability and expose explicit event and temporal attributes for trace consolidation and retrieval.

This representation is designed to make each memory unit independently retrievable while preserving the event and temporal context required for subsequent reasoning. It has two advantages: (1) \textbf{Multi-perspective event matching.} Explicit entity, location, and factual-time attributes characterize an event or state from complementary perspectives, helping distinguish answer-bearing memories from topically related events and improving their retrieval priority. (2) \textbf{Dual-time representation.} Separating factual time \(T_m^f\) from utterance time \(T_m^u\) makes temporal information directly comparable during retrieval. For example, if a user said on January 20, ``I went to the hospital yesterday,'' then \(T_m^u\) is January 20, while \(T_m^f\) is normalized to January 19. Resolving this distinction during memory construction supports temporal grounding and selection of the valid memory under time-sensitive or conflicting records.

\subsubsection{Dynamic Memory Trace Consolidation}
Atomic memory units preserve fine-grained evidence, but long-term conversations often contain state changes that cannot be captured by isolated units alone. For example, a user may first set a goal, later revise it, and finally ask about the previous goal. To make such update trajectories explicit, \method{} consolidates related memory units into dynamic memory traces. Write the structured fact of unit \(m\) as \(f_m=(a_m,p_m,o_m)\), where \(a_m\), \(p_m\), and \(o_m\) are its subject, predicate, and object, respectively. With the normalized text encoder \(E(\cdot)\), we first compute three pairwise similarities and then combine them into a relation score:
\begin{equation}
\begin{aligned}
s_f^{ij} &= \cos\!\left(E(f_{m_i}),E(f_{m_j})\right),\\
s_a^{ij} &= \cos\!\left(E(a_{m_i}),E(a_{m_j})\right),\\
s_p^{ij} &= \cos\!\left(E(p_{m_i}),E(p_{m_j})\right),\\
S_{\mathrm{rel}}(m_i,m_j) &=
\min\!\left\{s_f^{ij},s_a^{ij},s_p^{ij}\right\}.
\end{aligned}
\end{equation}
Here, \(s_a^{ij}\) and \(s_p^{ij}\) ensure that two facts describe the same state holder and relation, while \(s_f^{ij}\) rejects otherwise unrelated assertions without requiring identical objects. We take their minimum because a false merge can corrupt an entire trajectory, whereas unmerged units remain independently retrievable. For each unit, we test its 20 nearest fact-level neighbors. A set of memory units forms a relation cluster only if the relation score between every pair of their structured facts is no smaller than \(\tau_r\). We set \(\tau_r=0.80\) based on the sensitivity analysis in Figure~\ref{fig:hyperparameter-sensitivity}.

For the \(h\)-th cluster, \(\psi_h\) denotes its representative subject--predicate relation. An LLM-based consolidator orders the facts chronologically, merges repeated or semantically equivalent states, and retains distinct state changes. It treats a later incompatible value as an \emph{update}, a compatible refinement as a \emph{supplement}, and mutually compatible values as \emph{coexisting states}. The resulting trace is \(r_h=(\psi_h,H_h,s_h)\), where \(H_h\) is the ordered state trajectory and \(s_h\) is its latest supported state:
\begin{equation}
r_h=(\psi_h, H_h, s_h).
\end{equation}
If consolidation produces \(N_R\) traces, the resulting trace set is
\begin{equation}
\mathcal{R}=\{r_h\}_{h=1}^{N_R}.
\end{equation}
At query time, each retrieved atomic unit remains linked to its dynamic trace. The trace supplies the state trajectory to the reranker and generator, while the latest member provides a current-state representative when several retrieved units describe different versions of the same relation. This separation preserves source-level granularity for verification and retrieval, while linking distant updates into a coherent trajectory for conflict resolution and multi-hop evidence composition.

\begingroup
\setlength{\parskip}{0pt}
\fixedsubsection{Query-Aware Working Memory Activation}

\method{} activates working memory by selecting a compact set of query-relevant memory units from the long-term memory store \(\mathcal{S}=(\mathcal{M},\mathcal{R})\). The procedure follows two successive steps: hybrid memory retrieval and query-aware reranking. Hybrid memory retrieval identifies complementary candidates through lexical, semantic, and event-attribute retrieval. Lexical retrieval preserves exact query cues, semantic retrieval identifies related expressions, and event-attribute retrieval distinguishes memories with similar content but different entities, locations, or temporal conditions. RRF~\citep{cormack2009rrf} combines the three ranked lists according to rank positions, producing a unified candidate ordering without cross-channel score calibration. Query-aware reranking subsequently evaluates the fused candidates and selects the final working memory under a fixed budget, prioritizing memories that are relevant and consistent with the current state. Algorithm~\ref{alg:working-memory-activation} summarizes the complete procedure.

\fixedsubsubsection{Hybrid Memory Retrieval}
Given a query \(q\) issued at time \(t_q\), QGMem retrieves candidate memory units \(m\in\mathcal{M}\) through three complementary channels. Lexical retrieval, implemented with BM25~\citep{robertson2009bm25}, identifies units containing exact query cues such as names, dates, and locations, whereas semantic retrieval identifies related units expressed through aliases or paraphrases using dense embeddings. For event-attribute retrieval, the parser \(\phi(q,t_q)\) extracts the entity, location, and temporal constraint sets \(\mathcal{C}^{\mathrm{ent}}_q\), \(\mathcal{C}^{\mathrm{loc}}_q\), and \(\mathcal{C}^{\mathrm{time}}_q\). \(\operatorname{SemMatch}\) evaluates the compatibility of entity and location constraints with \(\mathcal{E}_m\) and \(\mathcal{P}_m\), while \(\operatorname{TimeMatch}\) evaluates the temporal constraint against \((T_m^f,T_m^u)\). Writing \(\mathcal{C}^{x}_q\) for the corresponding constraint set, the available scores are averaged over nonempty constraints to obtain \(S_{\mathrm{ELT}}\).

Each channel retains its \(n\) highest-scoring units and produces a ranked list \(\mathcal{L}_b\), \(\mathcal{L}_d\), or \(\mathcal{L}_e\). We use \(\operatorname{TopK}^{b}_{x\in\mathcal{X}}s(x)\) to denote the \(b\) highest-scoring elements of \(\mathcal{X}\) under score \(s\). RRF~\citep{cormack2009rrf} assigns the fused score \(S_{\mathrm{RRF}}(m,q)\) according to \(\operatorname{rank}_{\mathcal{L}}(m)\), the position of \(m\) in list \(\mathcal{L}\). Its rank-offset constant \(\kappa\) is fixed to 60. The \(n\) highest-scoring units after fusion form the reranking candidate pool \(\mathcal{C}_0\). This fusion preserves complementary retrieval coverage without requiring score calibration, particularly when long histories contain many superficially similar memories.

\fixedsubsubsection{Query-Aware Reranking}
Candidates in \(\mathcal{C}_0\) still differ in content completeness and answer relevance because first-stage scores consider each signal separately. \method{} therefore constructs a reranking document for every candidate containing its context-independent fact summary, source dialogue, entities, locations, factual time, utterance time, and the state trajectory from its associated dynamic memory trace when available. \(\operatorname{Reranker}(q,m)\), implemented with Qwen3-Reranker-8B, jointly scores the query and this structured document. Under the working-memory budget \(k\), the highest-scoring candidates form the activated working memory \(\mathcal{M}_q\). Its ordering is used by both graph construction and answer generation. This two-stage activation preserves candidate-stage coverage while concentrating complete, current, and answer-relevant memories at the highest ranks. The bounded working context retains the evidence needed for generation with controlled input size, supporting retrieval quality, context efficiency, and stable access as the memory store grows.

\endgroup

\tightsubsection{Graph-Augmented Generation}

The graph token encodes relational topology, while temporal order and state validity remain in the dynamic traces and serialized working memory. These complementary inputs support source-grounded, relation-aware generation without constructing a graph over the full history.

\subsubsection{Local Memory Graph Construction}
Each structured fact \(f_m=(u,\rho,v)\) in the activated working memory \(\mathcal{M}_q\) is represented by an entity node for the subject \(u\), a value node for the object \(v\), and a directed edge labeled by the predicate \(\rho\). The mappings \(\nu_{\mathrm{ent}}(\cdot)\) and \(\nu_{\mathrm{val}}(\cdot)\) normalize entity and value labels, respectively, such that their repeated occurrences share the same node. The induced node and directed-edge sets are
\begin{equation}
\begin{aligned}
\mathcal{V}_q
&=\bigcup_{m\in\mathcal{M}_q}\ 
  \{\nu_{\mathrm{ent}}(u_m),\nu_{\mathrm{val}}(v_m)\},\\
\mathcal{A}_q
&=\left\{
  (\nu_{\mathrm{ent}}(u_m),\rho_m,\nu_{\mathrm{val}}(v_m))
  \;\middle|\;
  \substack{m\in\mathcal{M}_q,\\[-0.2ex]
  f_m=(u_m,\rho_m,v_m)}
  \right\}.
\end{aligned}
\label{eq:local-memory-graph}
\end{equation}
The resulting local memory graph is \(\mathcal{G}_q=(\mathcal{V}_q,\mathcal{A}_q)\). It encodes the relational topology of the activated facts, supporting relation-aware evidence composition. Time-aware retrieval and dynamic traces in the text branch preserve temporal order and state validity, complementing the graph topology during generation. Because \(\mathcal{G}_q\) is induced by the fixed-budget working memory \(\mathcal{M}_q\), its size remains bounded as the dialogue history grows.

\subsubsection{Graph Encoding}
QGMem initializes each node \(i\) and edge predicate \(\rho_e\) with the embeddings \(\mathbf{x}_i\) and \(\mathbf{r}_e\) produced by the fixed sentence encoder \(f_{\mathrm{emb}}\), which is shared with semantic retrieval.

To incorporate predicate information during message passing, the graph encoder applies \(L\) relation-aware TransformerConv layers~\citep{shi2021masked} with \(H\) attention heads. For the incoming edge set \(\mathcal{I}_q(i)=\{e=(j,\rho_e,i)\in\mathcal{A}_q\}\), the node states are initialized by \(\mathbf{h}_i^{(0)}=\mathbf{x}_i\). At layer \(\ell\) and head \(h\), the relation-aware update is
\begin{equation}
\begin{aligned}
\eta_{ie}^{(\ell,h)}
&=\frac{\bigl(\mathbf{W}_{Q}^{(\ell,h)}\mathbf{h}_i^{(\ell)}\bigr)^{\!\top}
\bigl(\mathbf{W}_{K}^{(\ell,h)}\mathbf{h}_j^{(\ell)}
+\mathbf{W}_{E}^{(\ell,h)}\mathbf{r}_{e}\bigr)}{\sqrt{d_h}},\\
\alpha_{ie}^{(\ell,h)}
&=\frac{\exp(\eta_{ie}^{(\ell,h)})}
{\sum_{e'\in\mathcal{I}_q(i)}\exp(\eta_{ie'}^{(\ell,h)})},\\
\mathbf{m}_{i}^{(\ell,h)}
&=\sum_{e=(j,\rho_e,i)\in\mathcal{I}_q(i)}\alpha_{ie}^{(\ell,h)}
\bigl(\mathbf{W}_{V}^{(\ell,h)}\mathbf{h}_j^{(\ell)}
+\mathbf{W}_{E}^{(\ell,h)}\mathbf{r}_{e}\bigr),\\
\mathbf{h}_i^{(\ell+1)}
&=\Gamma_{\ell}\!\left(
\mathbf{W}_{0}^{(\ell)}\mathbf{h}_i^{(\ell)}
+\mathop{\Vert}_{h=1}^{H}\mathbf{m}_{i}^{(\ell,h)}\right).
\end{aligned}
\label{eq:relation-aware-graph-layer}
\end{equation}
Here, \(d_h\) is the dimension of each attention head and all \(\mathbf{W}\) terms and \(\Gamma_{\ell}\) are learned transformations. The attention weights jointly depend on neighboring node states and predicate embeddings. Their aggregation therefore propagates both memory content and typed relations into the updated node representations. After \(L\) layers, global mean pooling summarizes the node representations as \(\mathbf{g}_q\), and a two-layer nonlinear projector maps it to the memory graph token \(\mathbf{z}_q\) in the LLM embedding space:

\begin{algorithm}[H]
\caption{Query-Aware Working Memory Activation}
\label{alg:working-memory-activation}
\scriptsize
\begin{algorithmic}[1]
\Require Memory set \(\mathcal{M}\); query \(q\) at \(t_q\); budgets \(n,k\); RRF constant \(\kappa\)
\Ensure Activated working memory \(\mathcal{M}_q\)
\State \((\mathcal{C}^{\mathrm{ent}}_q,\mathcal{C}^{\mathrm{loc}}_q,\mathcal{C}^{\mathrm{time}}_q) \gets \phi(q,t_q)\) \Comment{query event constraints}
\State \(\mathcal{L}_{b} \gets \operatorname{TopK}^{n}_{m\in\mathcal{M}}\mathrm{BM25}(q,m)\) \Comment{lexical retrieval}
\State \(\mathcal{L}_{d} \gets \operatorname{TopK}^{n}_{m\in\mathcal{M}}\mathrm{Dense}(q,m)\) \Comment{semantic retrieval}
\State \(\mathcal{L}_e\gets\emptyset\)
\If{\(\mathcal{C}^{\mathrm{ent}}_q\neq\emptyset\lor\mathcal{C}^{\mathrm{loc}}_q\neq\emptyset\lor\mathcal{C}^{\mathrm{time}}_q\neq\emptyset\)}
    \For{\(m\in\mathcal{M}\)}
        \If{\(\mathcal{C}^{\mathrm{ent}}_q\neq\emptyset\)}
            \State \(s_{\mathrm{ent}}(m,q)\gets\operatorname{SemMatch}(\mathcal{C}^{\mathrm{ent}}_q,\mathcal{E}_m)\)
        \EndIf
        \If{\(\mathcal{C}^{\mathrm{loc}}_q\neq\emptyset\)}
            \State \(s_{\mathrm{loc}}(m,q)\gets\operatorname{SemMatch}(\mathcal{C}^{\mathrm{loc}}_q,\mathcal{P}_m)\)
        \EndIf
        \If{\(\mathcal{C}^{\mathrm{time}}_q\neq\emptyset\)}
            \State \(s_{\mathrm{time}}(m,q)\gets\operatorname{TimeMatch}(\mathcal{C}^{\mathrm{time}}_q,(T^f_m,T^u_m))\)
        \EndIf
        \State \(S_{\mathrm{ELT}}(m,q)\gets
        \operatorname*{mean}_{\substack{x\in\{\mathrm{ent},\mathrm{loc},\mathrm{time}\}:\\\mathcal{C}^{x}_q\neq\emptyset}}s_x(m,q)\)
    \EndFor
    \State \(\mathcal{L}_{e}\gets\operatorname{TopK}^{n}_{m\in\mathcal{M}}S_{\mathrm{ELT}}(m,q)\) \Comment{event-attribute retrieval}
\EndIf
\State \(\mathcal{U}\gets\mathcal{L}_b\cup\mathcal{L}_d\cup\mathcal{L}_e\)
\For{\(m\in\mathcal{U}\)}
    \State \(S_{\mathrm{RRF}}(m,q)\gets
    \displaystyle\sum_{\substack{\mathcal{L}\in\{\mathcal{L}_b,\mathcal{L}_d,\mathcal{L}_e\}\\m\in\mathcal{L}}}
    \bigl(\kappa+\operatorname{rank}_{\mathcal{L}}(m)\bigr)^{-1}\)
\EndFor
\State \(\mathcal{C}_{0}\gets\operatorname{TopK}^{n}_{m\in\mathcal{U}}S_{\mathrm{RRF}}(m,q)\)
\State \(\mathcal{M}_{q}\gets\operatorname{TopK}^{k}_{m\in\mathcal{C}_0}\operatorname{Reranker}(q,m)\) \Comment{working memory}
\State \Return \(\mathcal{M}_q\)
\end{algorithmic}
\end{algorithm}
\begin{equation}
\begin{aligned}
\mathbf{g}_q
&=\frac{1}{|\mathcal{V}_q|}\sum_{i\in\mathcal{V}_q}\mathbf{h}_i^{(L)},\\
\mathbf{z}_q
&=\mathbf{W}_2\sigma\!\left(\mathbf{W}_1\mathbf{g}_q+\mathbf{b}_1\right)+\mathbf{b}_2.
\end{aligned}
\label{eq:graph-token-projection}
\end{equation}
The resulting \(\mathbf{z}_q\) is supplied to the LLM. The graph encoder and projector parameters are collected in \(\omega\), with their configuration reported in Section~\ref{app:training_settings} of the Appendix.

\subsubsection{Graph-Text Fusion}
The LLM text embedder converts the serialized working memory \(C_q\) into the textual token sequence \(\mathbf{h}_t\). As illustrated in Figure~\ref{fig:overview}, graph-text fusion concatenates the graph token \(\mathbf{z}_q\) with \(\mathbf{h}_t\) and supplies \([\mathbf{z}_q,\mathbf{h}_t]\) to the frozen LLM together with the query \(q\). For an answer sequence \(\mathbf{a}\), the resulting autoregressive distribution is \(p_{\theta,\omega}(a_j\mid\mathbf{a}_{<j},q,[\mathbf{z}_q,\mathbf{h}_t])\), where \(\theta\) and \(\omega\) are the frozen LLM parameters and trainable graph-module parameters, respectively. This fusion preserves readable, source-grounded memory content while supplying graph-structural signals for relation-aware generation.

\subsubsection{Training Objective}
\method{} first builds a persistent memory store through event-centric memory construction. Query-aware working memory activation and graph-augmented generation then transform query-relevant memories into textual and structural representations, which are supplied to an LLM for answer generation. These memory operations are decoupled from the LLM backbone model, allowing the framework to support different LLM backbone models through an appropriate generation interface. Accordingly, optimization is restricted to the parameters of the relation-aware graph encoder and the two-layer projector, whereas both the sentence encoder and the backbone LLM remain fixed throughout training. Given \(N_D\) graph-grounded training instances, the \(s\)-th instance comprises a query \(q^{(s)}\), serialized working memory \(C_q^{(s)}\), local graph \(\mathcal{G}_q^{(s)}\), and target sequence \(\mathbf{a}^{(s)}\). The graph branch produces \(\mathbf{z}_q^{(s)}\), while the text branch produces \(\mathbf{h}_t^{(s)}\); their concatenation conditions the frozen LLM together with \(q^{(s)}\). The objective is evaluated over the target tokens, whose total number is \(T_D=\sum_{s=1}^{N_D}|\mathbf{a}^{(s)}|\). The teacher-forced objective is
\begin{equation}
\begin{aligned}
\mathcal{J}(\omega)
&=-\frac{1}{T_D}\sum_{s=1}^{N_D}\sum_{j=1}^{|\mathbf{a}^{(s)}|}\\[-0.2ex]
&\quad \log p_{\theta,\omega}\!\left(
a_j^{(s)}\mid \mathbf{a}_{<j}^{(s)},q^{(s)},
[\mathbf{z}_q^{(s)},\mathbf{h}_t^{(s)}]\right).
\end{aligned}
\label{eq:graph-training-objective}
\end{equation}
We minimize \(\mathcal{J}(\omega)\) in \eqref{eq:graph-training-objective} with respect to \(\omega\). Consequently, answer supervision encourages the encoded graph representation to retain relational information that contributes to generation without altering the pretrained LLM. Detailed training settings are provided in Section~\ref{app:training_settings} of the Appendix.

\subsection{Backbone Adaptability and Efficiency Analysis}

\subsubsection{Adaptability Across LLM Backbones}
\label{sec:method-backbone-generalization}
\method{} supports different backbone models, enabling the complete framework to operate with both open-weight and proprietary LLMs without reliance on a specific model. For open-weight backbones, the projector maps the encoded local graph to a graph token in the embedding space of the target LLM, which is provided jointly with the textual working memory. For proprietary backbones accessed through an API, the embedding space is inaccessible; accordingly, the activated working memory and local graph are serialized as text. This interface-level adaptation preserves the memory construction, activation, and graph organization procedures across backbones while accommodating their respective access constraints.

\subsubsection{Efficiency Analysis}
Long-term memory construction is performed offline once, and the resulting atomic units, dynamic traces, and retrieval indexes are reused across queries. Atomic memory construction processes the dialogue through fixed-size overlapping windows. Therefore, the number of extraction operations grows linearly with the number of dialogue turns. These windows can be processed independently, and their construction cost is amortized over all subsequent queries. For dynamic trace consolidation, each atomic unit examines only its 20 nearest fact-level neighbors, avoiding exhaustive pairwise comparison over the complete memory store.

At query time, retrieval and reranking activate at most \(k\) atomic memory units. Since each activated unit contributes one primary structured fact, the induced local graph contains at most \(k\) directed edges and \(2k\) nodes before node deduplication. Local graph materialization therefore requires \(O(k)\) time and space. With the best working-memory budget \(k=10\), both graph construction and graph encoding operate on a bounded structure whose size does not grow with the complete dialogue history. This design confines explicit relational organization to query-relevant evidence while reusing the lightweight persistent memory across queries.

\section{Experiments}

We evaluate \method{} across complementary requirements for agent memory~\citep{tan2025membench}: accurate retrieval, conflict resolution, ultra-long-history scalability, backbone adaptability, and efficiency. We further analyze retrieval quality, component contributions, hyperparameters, and a representative case.

\vspace{-0.25\baselineskip}
\subsection{Experimental Setup}
\label{sec:exp-setup}
\vspace{0.25\baselineskip}

\subsubsection{Benchmarks}
For \textbf{accurate retrieval}, LoCoMo~\citep{maharana2024locomo} tests evidence retrieval and composition across multi-session dialogues, while LongMemEval-S~\citep{wu2025longmemeval} tests persistent recall, reasoning, and updates. For \textbf{conflict resolution}, FactConsolidation-SH and FactConsolidation-MH from MemoryAgentBench~\citep{hu2025memoryagentbench} cover single- and multi-hop selection among conflicting facts. For \textbf{ultra-long scalability}, BEAM~\citep{tavakoli2026beam} provides histories of 100K, 500K, 1M, and 10M tokens.

\subsubsection{Baselines}
We compare \method{} with seven representative systems. The \textbf{retrieval-augmented baseline} is RAG; the \textbf{flat-structured memory baselines} are Mem0~\citep{chhikara2025mem0} and LightMem~\citep{fang2026lightmem}; and the \textbf{graph-based memory baselines} are EMem-G~\citep{zhou2025emem}, LiCoMemory~\citep{huang2026licomemory}, MR-Agent~\citep{ji2026memoryreconstructed}, and AriadneMem~\citep{zhu2026ariadnemem}. We use the same questions and backbone model for all available systems on each benchmark. Further implementation details are provided in Section~\ref{app:exp_details} of the Appendix.

\subsubsection{Evaluation Metrics}
We use two answer-quality metrics. (1) \textbf{LLM-judged Score.} A judge evaluates whether each answer satisfies the reference or task rubric~\citep{zheng2023llmjudge}; we repeat judging 10 times and report mean and standard deviation. For BEAM, we follow its official protocol, scoring nine nugget-based abilities and Event Ordering with normalized Kendall's $\tau_b$, then macro-averaging the ten ability scores. (2) \textbf{Answer Semantic Coverage.} AlignScore-large~\citep{zha2023alignscore} measures coverage of reference information; uncertainty is estimated with 10,000 bootstrap resamples.

\begin{table*}[t]
\caption{Main results with Qwen3-8B across six long-term memory benchmarks. Boldface and underlining indicate the best and second-best performance respectively.}
\label{tab:main-results}
\centering
\footnotesize
\setlength{\tabcolsep}{3.5pt}
\renewcommand{\arraystretch}{0.98}
\begin{tabularx}{\textwidth}{@{}>{\centering\arraybackslash}p{0.14\textwidth}|*{2}{>{\centering\arraybackslash}X}|*{2}{>{\centering\arraybackslash}X}|*{2}{>{\centering\arraybackslash}X}@{}}
\toprule
\multirow{3}{*}{\textbf{Method}}
& \multicolumn{2}{c|}{\textit{Accurate Retrieval}}
& \multicolumn{2}{c|}{\textit{Conflict Resolution}}
& \multicolumn{2}{c}{\textit{Ultra-Long Scalability}} \\
\cmidrule(lr){2-3}\cmidrule(lr){4-5}\cmidrule(lr){6-7}
& \textbf{LoCoMo} & \textbf{LongMemEval-S} & \textbf{Fact-SH} & \textbf{Fact-MH} & \textbf{BEAM-100K} & \textbf{BEAM-500K} \\
& \textit{$N$=1540} & \textit{$N$=500} & \textit{$N$=400} & \textit{$N$=400} & \textit{$N$=400} & \textit{$N$=700} \\
\midrule
\multicolumn{7}{c}{\textbf{(a) LLM-judged Score (\%)}} \\
\midrule
\multicolumn{7}{l}{\textbf{\textit{Retrieval-Augmented Baseline}}} \\
RAG & $57.93\!\pm\!0.06$ & $34.74\!\pm\!0.10$ & $37.82\!\pm\!0.24$ & $3.93\!\pm\!0.17$ & $29.04$ & $28.00$ \\
\cmidrule(lr){1-7}
\multicolumn{7}{l}{\textbf{\textit{Flat-Structured Memory Baselines}}} \\
Mem0 & $42.18\!\pm\!0.18$ & $25.96\!\pm\!0.08$ & \underline{$75.92\!\pm\!0.17$} & $5.08\!\pm\!0.24$ & \underline{$31.61$} & \underline{$33.50$} \\
LightMem & $57.58\!\pm\!0.23$ & \underline{$56.42\!\pm\!0.18$} & $62.25\!\pm\!0.17$ & \underline{$6.50\!\pm\!0.00$} & $27.79$ & $29.75$ \\
\cmidrule(lr){1-7}
\multicolumn{7}{l}{\textbf{\textit{Graph-Based Memory Baselines}}} \\
EMem-G & $62.48\!\pm\!0.18$ & $49.00\!\pm\!0.16$ & $36.85\!\pm\!1.13$ & $4.30\!\pm\!0.16$ & $16.14$ & $16.13$ \\
LiCoMemory & \underline{$74.47\!\pm\!0.10$} & $47.20\!\pm\!0.14$ & $22.25\!\pm\!0.19$ & $1.50\!\pm\!0.12$ & $13.34$ & $13.03$ \\
MR-Agent & $43.46\!\pm\!0.15$ & $23.20\!\pm\!0.12$ & $20.50\!\pm\!0.21$ & $3.00\!\pm\!0.15$ & $12.89$ & $16.85$ \\
AriadneMem & $51.42\!\pm\!0.10$ & $11.00\!\pm\!0.09$ & $64.50\!\pm\!0.28$ & $4.75\!\pm\!0.18$ & $27.41$ & $28.70$ \\
\cmidrule(lr){1-7}
\rowcolor{methodrow}
\textbf{\method{} (ours)}
& $\mathbf{82.77\!\pm\!0.13}$
& $\mathbf{62.14\!\pm\!0.13}$
& $\mathbf{78.10\!\pm\!0.24}$
& $\mathbf{12.50\!\pm\!0.00}$
& $\mathbf{35.31}$
& $\mathbf{38.91}$ \\
\midrule
\multicolumn{7}{c}{\textbf{(b) Answer Semantic Coverage (\%)}} \\
\midrule
\multicolumn{7}{l}{\textbf{\textit{Retrieval-Augmented Baseline}}} \\
RAG & \underline{$43.54\!\pm\!1.14$} & $29.85\!\pm\!1.84$ & $38.71\!\pm\!2.39$ & $5.00\!\pm\!0.95$ & $7.32\!\pm\!1.00$ & $8.54\!\pm\!0.82$ \\
\cmidrule(lr){1-7}
\multicolumn{7}{l}{\textbf{\textit{Flat-Structured Memory Baselines}}} \\
Mem0 & $24.75\!\pm\!0.99$ & $26.54\!\pm\!1.64$ & $\mathbf{81.70\!\pm\!1.87}$ & $2.95\!\pm\!0.71$ & $8.82\!\pm\!1.14$ & \underline{$11.50\!\pm\!0.95$} \\
LightMem & $35.45\!\pm\!1.13$ & \underline{$48.53\!\pm\!1.81$} & $61.00\!\pm\!2.27$ & \underline{$7.65\!\pm\!1.16$} & \underline{$8.88\!\pm\!1.15$} & $9.50\!\pm\!0.86$ \\
\cmidrule(lr){1-7}
\multicolumn{7}{l}{\textbf{\textit{Graph-Based Memory Baselines}}} \\
EMem-G & $34.57\!\pm\!1.13$ & $18.37\!\pm\!1.60$ & $40.09\!\pm\!2.41$ & $7.57\!\pm\!1.21$ & $5.86\!\pm\!0.81$ & $5.98\!\pm\!0.57$ \\
LiCoMemory & $35.97\!\pm\!1.10$ & $38.30\!\pm\!1.96$ & $22.06\!\pm\!1.96$ & $2.33\!\pm\!0.57$ & $4.02\!\pm\!0.76$ & $4.20\!\pm\!0.57$ \\
MR-Agent & $20.44\!\pm\!0.91$ & $21.03\!\pm\!1.70$ & $20.39\!\pm\!1.88$ & $3.79\!\pm\!0.82$ & $3.98\!\pm\!0.66$ & $5.61\!\pm\!0.59$ \\
AriadneMem & $23.44\!\pm\!0.97$ & $6.93\!\pm\!0.93$ & $66.42\!\pm\!2.33$ & $5.71\!\pm\!1.02$ & $7.05\!\pm\!0.94$ & $10.07\!\pm\!0.90$ \\
\cmidrule(lr){1-7}
\rowcolor{methodrow}
\textbf{\method{} (ours)}
& $\mathbf{56.31\!\pm\!1.08}$
& $\mathbf{51.31\!\pm\!1.91}$
& \underline{$80.58\!\pm\!1.95$}
& $\mathbf{13.34\!\pm\!1.63}$
& $\mathbf{16.15\!\pm\!1.57}$
& $\mathbf{16.95\!\pm\!1.20}$ \\
\bottomrule
\end{tabularx}
\end{table*}

\vspace{-0.25\baselineskip}
\subsection{Quantitative Evaluation}
\label{sec:main-results}
\vspace{0.25\baselineskip}

\subsubsection{Accurate Retrieval}
Table~\ref{tab:main-results} shows that \method{} leads on LoCoMo and LongMemEval-S under both metrics. Its LLM-judged scores of 82.77\% and 62.14\% exceed the strongest baselines by 8.30 and 5.72 points; semantic coverage reaches 56.31\% and 51.31\%, gains of 12.77 and 2.78 points. Agreement across the two evaluators indicates that event-indexed storage and query-aware activation prioritize complete answer-bearing memories across multi-session evidence.

\subsubsection{Conflict Resolution}
On Fact-SH and Fact-MH, \method{} attains LLM-judged scores of 78.10\% and 12.50\%, surpassing the strongest baselines by 2.18 and 6.00 points. Fact-MH nearly doubles the strongest baseline, with semantic coverage improving by 5.69 points to 13.34\%; Fact-SH coverage reaches 80.58\%. The larger multi-hop gain supports the division of labor between dynamic traces and time-aware activation for state selection and the local graph for evidence composition.

\subsubsection{Ultra-Long Dialogue History Scalability}
Across both scales, \method{} achieves overall scores of 35.31\%/38.91\% on BEAM-100K/500K, exceeding Mem0 by 3.70/5.41 points, and semantic coverage of 16.15\%/16.95\%, with gains of 7.27/5.45 points. Its nine-ability Nugget Scores are 36.81\%/40.26\%, with Event Ordering $\tau_b$ of 21.77\%/26.79\%, showing that compact activation supports diverse memory abilities under long histories.

Figure~\ref{fig:beam-scalability} extends BEAM from 100K to 10M tokens; methods share questions and histories within each scale, while question sets differ across scales. \method{} leads throughout, remains above 35\% through 1M tokens, and scores 29.81\% at 10M, exceeding LightMem and AriadneMem by 20.18 and 21.21 points. Event-indexed retrieval limits irrelevant-history interference, reranking selects a fixed-budget working memory, and the local graph grows with activated evidence rather than full history.

\begin{figure}[!t]
\centering
\includegraphics[width=0.88\linewidth]{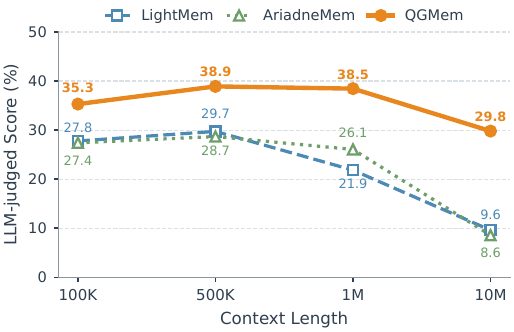}
\compactfigurecaption{fig:beam-scalability}{Scalability on BEAM with Qwen3-8B. All methods use the same questions and histories within each scale, while the evaluation sets differ across scales.}
\end{figure}

\subsubsection{Retrieval Quality Analysis}
Recall@10 measures gold-evidence coverage, Precision@10 the concentration of relevant units, Sufficiency@10 whether the set jointly supports the answer, and nDCG@10 rank quality~\citep{jarvelin2002ndcg}. Table~\ref{tab:evidence-retrieval} shows that \method{} combines 89.08\% Recall@10 with the best Precision@10 (20.92\%), Sufficiency@10 (97.11\%), and nDCG@10 (82.50\%). Precision and nDCG improve by 3.00 and 14.34 points, indicating that answer-supporting memories are both covered and concentrated at high ranks; the sufficiency score confirms coverage of combined multi-evidence requirements.

\begin{table}[!t]
\centering
\footnotesize
\compacttablecaption{tab:evidence-retrieval}{The evidence retrieval performance of \method{} compared with representative baselines. Higher values indicate better performance for all metrics.}
\renewcommand{\arraystretch}{0.98}
\begin{tabularx}{\linewidth}{@{}l *{4}{>{\centering\arraybackslash}X}@{}}
\toprule
\textbf{Method} & \textbf{R@10} & \textbf{P@10} & \textbf{Suff.@10} & \textbf{nDCG@10} \\
\midrule
RAG & \textbf{90.34} & \underline{17.92} & {96.71} & \underline{68.16} \\
Mem0 & 82.77 & 16.42 & 86.71 & 48.13 \\
LightMem & 80.67 & 11.04 & \underline{89.60} & 43.46 \\
EMem-G & 78.57 & 5.09 & 80.35 & 26.79 \\
LiCoMemory & 81.51 & 12.54 & 84.97 & 47.46 \\
MR-Agent & 55.88 & 3.99 & 45.09 & 28.73 \\
AriadneMem & 55.46 & 8.55 & 56.07 & 28.56 \\
\rowcolor{methodrow}
\textbf{\method{} (ours)} & \underline{89.08} & \textbf{20.92} & \textbf{97.11} & \textbf{82.50} \\
\bottomrule
\end{tabularx}
\par\vspace{0.7\baselineskip}
\compacttablecaption{tab:efficiency}{Efficiency on LoCoMo with Qwen3-8B. Online LLM tokens include all query-time calls; offline latency is amortized per QA.}
\renewcommand{\arraystretch}{0.98}
\setlength{\tabcolsep}{2.5pt}
\begin{tabularx}{\linewidth}{@{}l*{4}{>{\centering\arraybackslash}X}@{}}
\toprule
\textbf{Method} & \multicolumn{2}{c}{\textbf{Online}} & \multicolumn{1}{c}{\textbf{Ret. ctx.}} & \multicolumn{1}{c}{\textbf{Offline}} \\
\cmidrule(lr){2-3}\cmidrule(lr){4-4}\cmidrule(l){5-5}
& \textbf{tok./QA} & \textbf{Lat. (s/QA)} & \textbf{tok./QA} & \textbf{amort. s/QA} \\
\midrule
RAG & 3430.01 & 3.84 & 3194.50 & 0.02 \\
Mem0 & 879.68 & 3.49 & 656.12 & 5.07 \\
LightMem & 819.29 & 2.99 & 620.79 & 1.48 \\
LiCoMemory & 616.48 & 5.13 & 391.95 & 0.57 \\
EMem-G & 1505.78 & 7.18 & 563.35 & 0.24 \\
\rowcolor{methodrow}
\textbf{\method{} (ours)} & 1208.79 & 5.94 & 598.19 & 1.09 \\
\bottomrule
\end{tabularx}
\end{table}

\subsubsection{Adaptability Across LLM Backbones}
We compare Qwen3-8B, Llama3.1-8B-Instruct~\citep{grattafiori2024llama3}, and GPT-4o-mini~\citep{openai2024gpt4omini} on four benchmarks. Figure~\ref{fig:backbone-generalization} shows the highest average for every backbone: 64.58\%, 56.00\%, and 70.11\%, exceeding the strongest corresponding baselines by 13.57, 5.31, and 13.80 points.

\begin{figure*}[!t]
\centering
\includegraphics[width=0.89\textwidth]{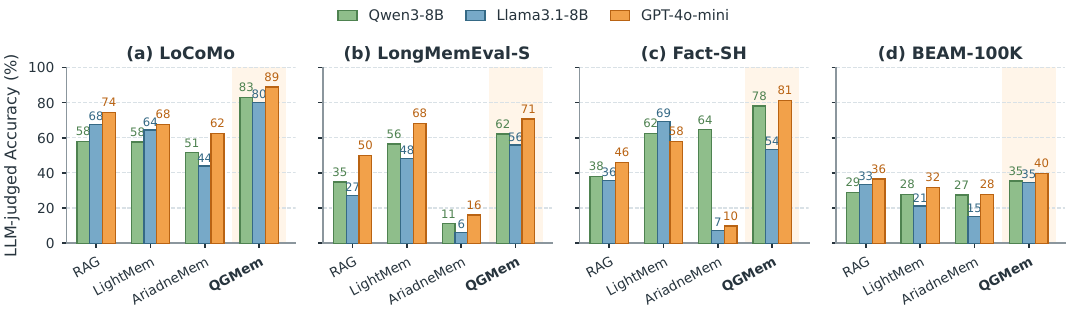}
\compactfigurecaption{fig:backbone-generalization}{Adaptability across LLM backbones on four benchmarks with Qwen3-8B, Llama3.1-8B-Instruct, and GPT-4o-mini.}
\end{figure*}

\method{} leads on LoCoMo, LongMemEval-S, and BEAM-100K for all three backbones, and reaches 78.10\%/81.25\% on Fact-SH with Qwen3-8B/GPT-4o-mini. GPT-4o-mini receives a textual serialization because its embedding space is inaccessible; its gains show that event-indexed memory and query-aware organization transfer across graph-token and text-only interfaces.

\subsubsection{Efficiency Analysis}
Table~\ref{tab:efficiency} reports online LLM and retrieved-context tokens, end-to-end online latency, and amortized offline construction latency on LoCoMo. \method{} uses 598.19 retrieved-context and 1208.79 total online tokens per QA, completes the full retrieval-to-generation pipeline in 5.94 seconds, and improves the LoCoMo score over EMem-G by 20.29 points. The 1.09-second offline value amortizes one-time construction of atomic units, traces, and indexes across all questions. Because the local graph is induced from fixed-budget working memory, longer histories enlarge the retrieval store without increasing the maximum graph supplied to the encoder or generator.

\begin{table*}[!t]
\centering
\compacttablecaption{tab:component-ablation}{Component ablation of \method{} with Qwen3-8B on each benchmark. Macro Avg. is the unweighted mean across the six benchmarks, assigning equal weight to each benchmark. Higher values are better; the best and second-best results are \textbf{bolded} and \underline{underlined} respectively.}
\footnotesize
\setlength{\tabcolsep}{4pt}
\renewcommand{\arraystretch}{0.98}
\begin{tabularx}{\textwidth}{@{}l|*{2}{>{\centering\arraybackslash}X}|*{2}{>{\centering\arraybackslash}X}|*{2}{>{\centering\arraybackslash}X}|>{\centering\arraybackslash}X@{}}
\toprule
\textbf{Setting} & \textbf{LoCoMo} & \mbox{\scriptsize\textbf{LongMemEval-S}} & \textbf{Fact-SH} & \textbf{Fact-MH} & \textbf{BEAM-100K} & \textbf{BEAM-500K} & \mbox{\scriptsize\textbf{Macro Avg.}} \\
\midrule
\rowcolor{methodrow}
\textbf{Full \method{}} & \textbf{82.77} & \textbf{62.14} & \textbf{78.10} & \textbf{12.50} & \underline{35.31} & \textbf{38.91} & \textbf{51.62} \\
\midrule
\rowcolor{tablegroup}
\multicolumn{8}{l}{\textbf{\textit{Memory Construction Phase}}} \\
\textit{w/o} Dynamic Memory Traces & 81.09 & 61.29 & 75.82 & 9.61 & 34.78 & 34.68 & 49.55$^{\textcolor{red}{\downarrow 2.08}}$ \\
\textit{w/o} Atomic Memory Units & 79.84 & 48.02 & 71.75 & 7.08 & 27.28 & 31.16 & 44.19$^{\textcolor{red}{\downarrow 7.43}}$ \\
\midrule
\rowcolor{tablegroup}
\multicolumn{8}{l}{\textbf{\textit{Memory Retrieval Phase}}} \\
\textit{w/o} Hybrid memory retrieval & \underline{81.93} & \underline{61.58} & 75.32 & \underline{11.90} & 33.01 & \underline{37.34} & \underline{50.18}$^{\textcolor{red}{\downarrow 1.44}}$ \\
\textit{w/o} Reranker & 73.82 & 52.56 & \underline{76.91} & 8.88 & 32.78 & 33.38 & 46.39$^{\textcolor{red}{\downarrow 5.23}}$ \\
\midrule
\rowcolor{tablegroup}
\multicolumn{8}{l}{\textbf{\textit{Graph-Augmented Generation Phase}}} \\
\textit{w/o} Graph Token & 81.80 & 55.08 & 70.29 & 9.10 & \textbf{35.75} & 34.00 & 47.67$^{\textcolor{red}{\downarrow 3.95}}$ \\
\bottomrule
\end{tabularx}
\end{table*}

\subsection{Ablation and Analysis}
\label{sec:ablation-analysis}
\subsubsection{Component ablation}
Table~\ref{tab:component-ablation} reports five variants; Macro Avg. is the unweighted mean across six benchmarks. For memory construction, excluding atomic memory units produces the largest decrease in Macro Avg. at 7.43 points. The decreases are 14.12 points on LongMemEval-S, 8.03/7.75 on BEAM-100K/500K, and 5.42 on Fact-MH. Event-level retrieval granularity therefore affects both long-history retrieval and multi-fact composition. Dynamic memory traces contribute 2.08 points to Macro Avg., with larger effects on Fact-MH (2.89 points) and BEAM-500K (4.23 points), two settings that place greater demands on tracking distant state changes.

For memory retrieval, excluding the reranker decreases Macro Avg. by 5.23 points, the second-largest change in the table. The corresponding decreases are 8.95/9.58 points on LoCoMo/LongMemEval-S and 2.53/5.53 points on BEAM-100K/500K. Candidate fusion therefore requires query-specific prioritization before answer generation. Excluding hybrid retrieval has a smaller overall effect of 1.44 points, but decreases Fact-SH and BEAM-100K by 2.78 and 2.30 points, respectively. The three retrieval channels thus improve candidate coverage before reranking.

For graph-augmented generation, excluding the graph token decreases Macro Avg. by 3.95 points. The decreases on Fact-SH, LongMemEval-S, Fact-MH, and BEAM-500K are 7.81, 7.06, 3.40, and 4.91 points, respectively. The reductions across these benchmarks suggest that graph topology complements the temporal and state-validity information carried by dynamic traces and serialized working memory.

\begin{figure*}[t]
\centering
\includegraphics[width=\textwidth]{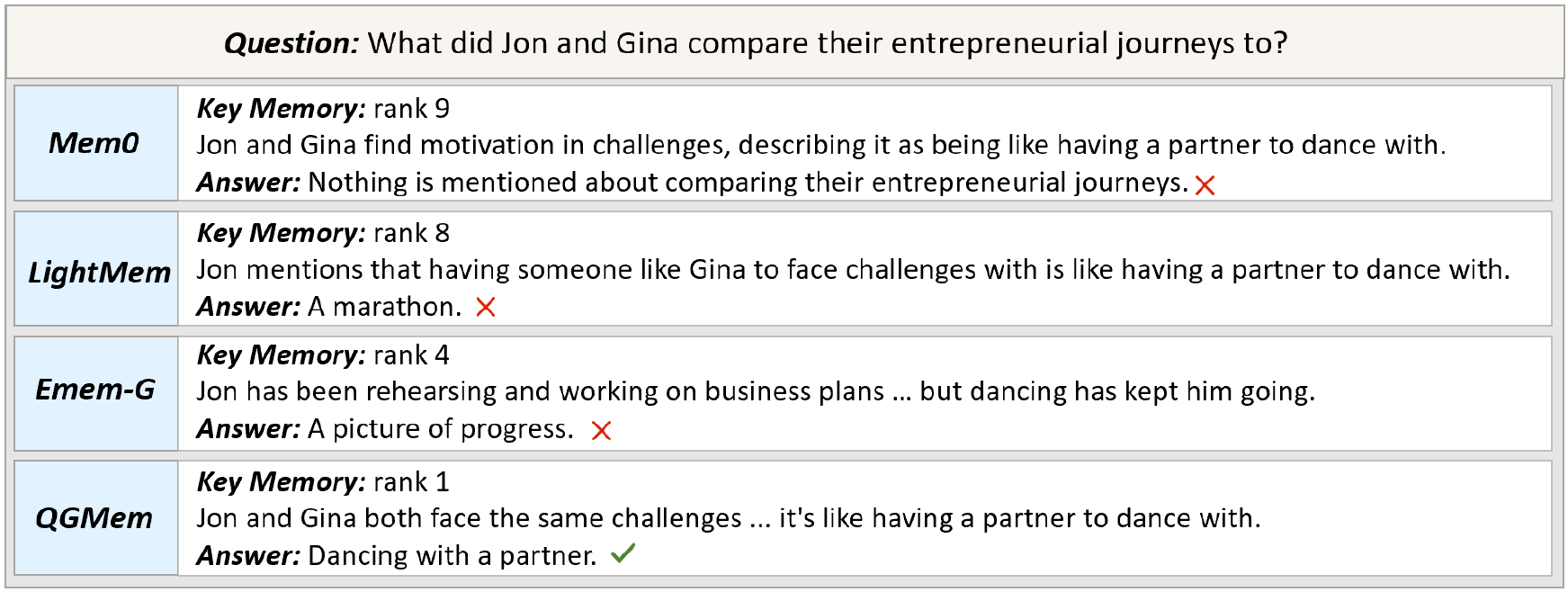}
\caption{A case study from LoCoMo. Mem0, LightMem, and EMem-G respectively illustrate ranking, evidence-utilization, and relation-selection failures, whereas \method{} promotes the answer-bearing comparison to Rank~1 and produces the grounded answer.}
\label{fig:case-study}
\vspace{0.7\baselineskip}

\includegraphics[width=\textwidth]{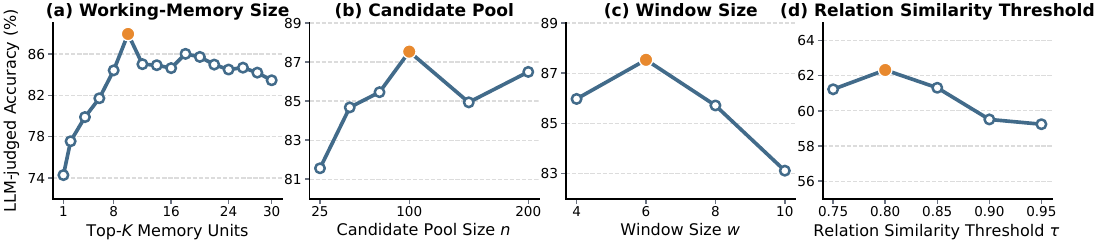}
\begingroup
\setlength{\abovecaptionskip}{-2pt}
\compactfigurecaption{fig:hyperparameter-sensitivity}{Hyperparameter sensitivity of \method{} with Qwen3-8B. Panels (a) through (d) vary working memory size, candidate-pool size, construction-window size, and relation similarity threshold respectively. Orange markers indicate the maximum on each curve.}
\endgroup
\end{figure*}

\subsubsection{Qualitative analysis}
Figure~\ref{fig:case-study} illustrates three failures: Mem0 retrieves ``a partner to dance with'' only at Rank~9; LightMem retrieves it at Rank~8 but answers ``a marathon''; and EMem-G selects a nearby motivation relation at Rank~4 and answers ``a picture of progress.'' These correspond to ranking, evidence-utilization, and relation-selection errors.

\method{} preserves the comparison as a traceable atomic unit, reranks it to Rank~1, and connects Jon and Gina's shared challenges to the comparison in the local graph. Text and graph evidence therefore agree on ``dancing with a partner,'' illustrating how memory granularity, selective activation, and local relational organization address the three failure points.

\subsubsection{Hyperparameter sensitivity}
Figure~\ref{fig:hyperparameter-sensitivity} varies one parameter at a time. For the retrieval budgets, accuracy increases with working-memory size up to \(K{=}10\) and remains below this maximum at every larger value evaluated. The initial increase is consistent with improved evidence coverage, while the subsequent reduction indicates that additional units can introduce irrelevant or redundant evidence. Candidate-pool size has a different trend. Accuracy is highest at \(n{=}100\), and larger pools provide no consistent improvement, indicating that the reranker has sufficient candidate diversity at this value.

For memory construction, \(w{=}6\) achieves the highest accuracy by providing sufficient context for local references while retaining event-level separation. The neighboring settings remain competitive, although performance decreases when the window grows to ten turns. The macro-averaged result for relation consolidation is highest at \(\tau{=}0.80\). Lower thresholds retain more weakly related facts, while higher thresholds remove useful cross-event connections. The observed values are consistent with a trade-off between relation coverage and relevance.

\FloatBarrier
\section{Conclusion}
\method{} is an event-centric conversational memory framework with query-aware graph augmentation. Its atomic units and dynamic traces preserve grounded evidence and evolving states, while query-aware activation organizes relevant units for local graph reasoning. Across six benchmarks and contexts up to 10M tokens, \method{} improves retrieval, multi-hop reasoning, conflict resolution, and scalability across diverse backbones with compact contexts.

\bibliographystyle{IEEEtran}
\bibliography{custom}

\clearpage
\appendix
\raggedbottom
\input{appendix}

\end{document}

%% file: appendix.tex
\section*{I. Implementation Details}
\label{app:implementation_details}

\subsection{Implementation Settings}

\subsubsection{Memory construction}
In the stage of memory construction, dialogue histories are first split at original session boundaries and then scanned within each session using fixed-size sliding windows. Unless otherwise specified, we use a window size of 6 turns and an overlap of 2 turns, resulting in a step size of 4 turns. Overlap turns are explicitly marked in the extraction prompt and are used only as background context for resolving pronouns and local references; the extractor is instructed not to create memory units whose evidence lies only in the overlap. The extractor may generate multiple memory units from the same window when different facts, preferences, events, or state changes are independently retrievable. BGE-base-en-v1.5~\citep{xiao2024cpack} is then used to embed memory-unit labels and graph records for retrieval and graph construction.

\subsubsection{Retrieval and generation}
For all LLM-driven steps, we set the decoding temperature to 0. During retrieval, the candidate pool size is set to 100 after RRF fusion over lexical, semantic, and event-attribute retrieval results, and reranking selects the best-performing budget of 10 memory units as working memory. For entity and location matching in the event-attribute retriever, dense semantic matching uses a similarity threshold of 0.60. When memory evolution is enabled, relation clustering uses a relation similarity threshold of 0.80 and a minimum relation-cluster size of 2.

\subsection{Training Settings}
\label{app:training_settings}

Training data are constructed from ConvoMem~\citep{pakhomov2025convomem}, a conversational memory benchmark built from multi-turn dialogue histories. We draw examples from three deterministic-answer categories: {user\_evidence}, {assistant\_facts\_evidence}, and {changing\_evidence}, which respectively test user fact recall, assistant fact recall, and updated information tracking. These categories cover memory capabilities aligned with LoCoMo and LongMemEval-S, such as user/assistant information recall, temporal grounding, and knowledge update. We sample the three categories using a fixed 5:3:4 ratio, yielding 1,200 cases and 2,138 QA pairs.

For each case, we run the full QGMem pipeline and record the trajectory information, including the activated memory units, their serialized textual content, the local memory graph, the generated answer, and the reasoning path. We retain only cases with correct final answers, resulting in 1,634 judge-correct graph-grounded QA examples. We split the retained examples into train/validation/test partitions, and checkpoint selection is based only on the validation split. No LoCoMo or LongMemEval-S examples are used in this part. The retained ConvoMem examples and their trajectory information are used as supervision for training the memory graph token. Node and edge labels are represented with 768-dimensional BGE-base-en-v1.5 embeddings. The target is a JSON sequence containing \texttt{answer} and \texttt{reasoning\_path} fields.

\begin{table}[t]
\caption{Training hyperparameters for the memory graph token.}
\label{tab:graph-token-training}
\centering
\footnotesize
\setlength{\tabcolsep}{5pt}
\renewcommand{\arraystretch}{0.98}
\begin{tabular}{@{}lc@{}}
\toprule
\textbf{Hyperparameter} & \textbf{Value} \\
\midrule
Backbone LLM & Qwen3-8B, Llama3.1-8B-Instruct \\
LoRA & Disabled \\
Trainable modules & Graph encoder, projector \\
Graph encoder & 4 TransformerConv layers \\
Hidden size & 768 \\
Attention heads & 4 \\
Pooling & Mean pooling \\
Projector & \(768 \rightarrow 2048 \rightarrow 4096\) \\
Batch size & 4 \\
Gradient accumulation & 4 \\
Learning rate & \(1\times10^{-4}\) \\
Weight decay & 0.05 \\
Training epochs & 10 \\
Warmup epochs & 1 \\
\bottomrule
\end{tabular}
\end{table}

\subsection{Evidence Retrieval Quality Evaluation Protocol}
\label{app:evidence-retrieval}

Since different memory systems use different memory granularities, we evaluate retrieval quality at the gold-evidence level rather than by exact text overlap. We use the full LoCoMo evaluation set under the Qwen3-8B setting and compare \method{} with the complete LoCoMo baseline suite: RAG, Mem0~\citep{chhikara2025mem0}, LightMem~\citep{fang2026lightmem}, EMem-G~\citep{zhou2025emem}, LiCoMemory~\citep{huang2026licomemory}, MR-Agent~\citep{ji2026memoryreconstructed}, and AriadneMem~\citep{zhu2026ariadnemem}. All systems use the full evaluation set and the same retrieval cutoff \(\ell=10\). For each question, let \(\mathcal{E}^{\star}\) denote the gold-evidence set and \(\widehat{\mathcal{M}}_{\ell}\) denote the top-\(\ell\) retrieved memory units. A GPT-4o-mini judge identifies the covered subset \(\operatorname{Covered}(\widehat{\mathcal{M}}_{\ell},\mathcal{E}^{\star})\subseteq\mathcal{E}^{\star}\), the relevant subset \(\operatorname{Relevant}(\widehat{\mathcal{M}}_{\ell},\mathcal{E}^{\star})\subseteq\widehat{\mathcal{M}}_{\ell}\), and whether the retrieved set contains sufficient answer-supporting evidence. We compute:
\begin{equation}
\mathrm{Recall@}\ell=
\frac{|\operatorname{Covered}(\widehat{\mathcal{M}}_{\ell},\mathcal{E}^{\star})|}
{|\mathcal{E}^{\star}|},
\end{equation}
\begin{equation}
\mathrm{Precision@}\ell=
\frac{|\operatorname{Relevant}(\widehat{\mathcal{M}}_{\ell},\mathcal{E}^{\star})|}
{\ell},
\end{equation}
and \(\mathrm{Sufficiency@}\ell=1\) if the retrieved set as a whole contains sufficient answer-supporting evidence to answer the question, and \(0\) otherwise. Unlike a conventional hit indicator, which is satisfied by any one relevant retrieval, sufficiency requires the combined retrieved evidence to cover the information needed for the answer. To measure ranking quality, let \(y_i\in\{0,1\}\) indicate whether the memory unit at rank \(i\) is relevant, and let \(\mathrm{IDCG@}\ell\) denote the DCG of the ideal ordering for the same number of relevant retrieved memories. We compute
\begin{equation}
\mathrm{nDCG@}\ell=
\frac{1}{\mathrm{IDCG@}\ell}\sum_{i=1}^{\ell}
\frac{y_i}{\log_2(i+1)},
\end{equation}
Recall is micro-averaged over gold items. Precision@10 uses the fixed denominator \(10N\) across \(N\) questions, treating unfilled ranks as non-relevant, while Sufficiency@10 and nDCG@10 are averaged over questions. All scores in the evidence-retrieval table of the main paper are reported as percentages. Gold evidence and judge labels are used only for evaluation and are unavailable to all retrieval systems.

\subsection{Experiment Details}
\label{app:exp_details}

\subsubsection{Experimental Settings}
To ensure comparability, all systems use the same answer format, LLM-as-judge setting, and number of retrieved entries under each benchmark. Semantic retrieval uses BGE-base-en-v1.5~\citep{xiao2024cpack}, and reranking uses Qwen3-Reranker-8B~\citep{zhang2025qwen3embedding}. 

\subsection{Additional Ablation Studies}
\label{app:additional-ablation}
\subsubsection{Retrieval and Selection Ablation}
Table~\ref{tab:retrieval-step-ablation} ablates the retrieval and selection choices in working memory activation on LoCoMo with Qwen3-8B. Removing reranking leads to the largest drop, showing that semantic refinement after hybrid memory retrieval is important for selecting memory units containing answer-supporting evidence. Removing event-attribute retrieval also degrades performance, confirming that explicit entity, location, and time constraints help rank candidates matching the event attributes ahead of topically related but mismatched ones. Removing hybrid memory retrieval has a smaller but consistent effect, suggesting that the initial candidate pool improves recall before reranking.

\begin{table}[t]
\caption{Ablation of retrieval steps on LoCoMo with Qwen3-8B.}
\label{tab:retrieval-step-ablation}
\centering
\footnotesize
\setlength{\tabcolsep}{5pt}
\renewcommand{\arraystretch}{0.98}
\begin{tabularx}{\columnwidth}{@{}X>{\centering\arraybackslash}p{0.20\columnwidth}@{}}
\toprule
\textbf{Setting} & \textbf{Avg. Score} \\
\midrule
\rowcolor{methodrow}
\textbf{Full activation pipeline} & \textbf{82.77} \\
w/o Hybrid memory retrieval & 81.93 \\
w/o Event-attribute retrieval & 71.67 \\
w/o reranking & 73.82 \\
\bottomrule
\end{tabularx}
\end{table}
\section*{II. Prompts}
\label{app:prompts}
\subsection{LLM-AS-JUDGE}

\noindent\textit{1) LoCoMo Judge Prompt:}\par\nobreak
\begin{tcolorbox}[
  colback=gray!10,
  colframe=black,
  title={LoCoMo Judge Prompt},
  fonttitle=\bfseries\footnotesize,
  width=\columnwidth,
  left=1mm,
  right=1mm,
  top=1mm,
  bottom=1mm,
  boxrule=0.5pt,
    after skip=10pt
]
\footnotesize
You are an evaluator for long-term conversational question answering.

Given a question, a reference answer, and a model prediction, determine whether the prediction correctly answers the question.

A prediction should be judged as correct if it is semantically equivalent to the reference answer, even if the wording is different. Minor paraphrases are acceptable. A prediction should be judged as incorrect if it contradicts the reference answer, misses a key entity, event, location, time, or relation, gives an answer that is too vague, or contains unsupported information.

\textbf{Question:} \{question\}

\textbf{Reference answer:} \{reference\_answer\}

\textbf{Model prediction:} \{model\_prediction\}

Return only a JSON object with two fields: \texttt{\{"judgment": "correct" or "incorrect", "reason": "brief explanation"\}}
\end{tcolorbox}

\noindent\textit{2) LongMemEval-S Judge Prompt:}\par\nobreak

\begin{tcolorbox}[
  colback=gray!10,
  colframe=black,
  title={Single-session-user/assistant Multi-session},
  fonttitle=\bfseries\footnotesize,
  width=\columnwidth,
  left=1mm,
  right=1mm,
  top=1mm,
  bottom=1mm,
  boxrule=0.5pt,
  before skip=10pt,
    after skip=10pt
]
\footnotesize
I will give you a question, a correct answer, and a response from a model. Please answer yes if the response contains the correct answer. Otherwise, answer no. If the response is equivalent to the correct answer or contains all the intermediate steps to get the correct answer, you should also answer yes. If the response only contains a subset of the information required by the answer, answer no.

\textbf{Question:} \{question\}\\
\textbf{Correct Answer:} \{answer\}\\
\textbf{Model Response:} \{response\}

Is the model response correct? Answer yes or no only.
\end{tcolorbox}

\begin{tcolorbox}[
  colback=gray!10,
  colframe=black,
  title={Temporal Reasoning Tasks},
  fonttitle=\bfseries\footnotesize,
  width=\columnwidth,
  left=1mm,
  right=1mm,
  top=1mm,
  bottom=1mm,
  boxrule=0.5pt,
  before skip=10pt,
    after skip=10pt
]
\footnotesize
I will give you a question, a correct answer, and a response from a model. Please answer yes if the response contains the correct answer. Otherwise, answer no. If the response is equivalent to the correct answer or contains all the intermediate steps to get the correct answer, you should also answer yes. If the response only contains a subset of the information required by the answer, answer no. In addition, do not penalize off-by-one errors for the number of days. If the question asks for the number of days/weeks/months, etc., and the model makes off-by-one errors (e.g., predicting 19 days when the answer is 18), the model's response is still correct.

\textbf{Question:} \{question\}\\
\textbf{Correct Answer:} \{answer\}\\
\textbf{Model Response:} \{response\}

Is the model response correct? Answer yes or no only.
\end{tcolorbox}

\begin{tcolorbox}[
  colback=gray!10,
  colframe=black,
  title={Knowledge Update Tasks},
  fonttitle=\bfseries\footnotesize,
  width=\columnwidth,
  left=1mm,
  right=1mm,
  top=1mm,
  bottom=1mm,
  boxrule=0.5pt,
  before skip=10pt,
    after skip=10pt
]
\footnotesize
I will give you a question, a correct answer, and a response from a model. Please answer yes if the response contains the correct answer. Otherwise, answer no. If the response contains some previous information along with an updated answer, the response should be considered as correct as long as the updated answer is the required answer.

\textbf{Question:} \{question\}\\
\textbf{Correct Answer:} \{answer\}\\
\textbf{Model Response:} \{response\}

Is the model response correct? Answer yes or no only.
\end{tcolorbox}

\begin{tcolorbox}[
  colback=gray!10,
  colframe=black,
  title={Single-session Preference Tasks},
  fonttitle=\bfseries\footnotesize,
  width=\columnwidth,
  left=1mm,
  right=1mm,
  top=1mm,
  bottom=1mm,
  boxrule=0.5pt,
  before skip=10pt,
    after skip=10pt
]
\footnotesize
I will give you a question, a rubric for desired personalized response, and a response from a model. Please answer yes if the response satisfies the desired response. Otherwise, answer no. The model does not need to reflect all the points in the rubric. The response is correct as long as it recalls and utilizes the user's personal information correctly.

\textbf{Question:} \{question\}\\
\textbf{Rubric:} \{answer\}\\
\textbf{Model Response:} \{response\}

Is the model response correct? Answer yes or no only.
\end{tcolorbox}

\begin{tcolorbox}[
  colback=gray!10,
  colframe=black,
  title={Abstention Tasks},
  fonttitle=\bfseries\footnotesize,
  width=\columnwidth,
  left=1mm,
  right=1mm,
  top=1mm,
  bottom=1mm,
  boxrule=0.5pt,
  before skip=10pt,
    after skip=10pt
]
\footnotesize
I will give you an unanswerable question, an explanation, and a response from a model. Please answer yes if the model correctly identifies the question as unanswerable. The model could say that the information is incomplete, or some other information is given but the asked information is not.

\textbf{Question:} \{question\}\\
\textbf{Explanation:} \{answer\}\\
\textbf{Model Response:} \{response\}

Does the model correctly identify the question as unanswerable? Answer yes or no only.
\end{tcolorbox}

\input{appendix_prompts_latest}

%% file: appendix_prompts_latest.tex
\subsubsection{Prompts for QGMem}
The following templates follow the current implementation. Atomic memory construction has two LLM stages. The first stage segments an annotated sliding window and returns one source-grounded claim and one canonical triple per memory unit. The second stage receives those memory units in a batch and returns retrieval indexes keyed by the unchanged \texttt{memory\_unit\_id}. Event and activity terms are included in entities; the current implementation does not use a separate event-index field. The final-answer variable \texttt{\{serialized\_graph\}} contains the complete textual generation context: evidence clusters with labeled memory-unit text and trajectories, followed by labeled structured facts. For open-weight backbones, the learned graph token is injected at the embedding level and therefore has no textual prompt placeholder.

\begin{tcolorbox}[
  enhanced,
  breakable,
  colback=gray!10,
  colframe=black,
  title={Atomic Memory Unit Extraction Prompt},
  fonttitle=\bfseries\footnotesize,
  width=\columnwidth,
  left=1mm,
  right=1mm,
  top=1mm,
  bottom=1mm,
  boxrule=0.5pt,
  before skip=6pt,
  after skip=6pt
]
\begin{lstlisting}[style=promptstyle,basicstyle=\rmfamily\footnotesize]
[System]
You are a long-conversation memory-unit extractor.

# Input
- MEMORY_CONTENT contains a fixed-size sliding window of dialogue lines in this form:
  NEW_IN_THIS_WINDOW [source_id] time: <timestamp> speaker: <message>
  OVERLAP_FROM_PREVIOUS_WINDOW [source_id] time: <timestamp> speaker: <message>
- OVERLAP_FROM_PREVIOUS_WINDOW lines were already included in the previous sliding window. Do not extract memory units from these turns; they serve only as background information for resolving references and ambiguity for NEW_IN_THIS_WINDOW turns.
- NEW_IN_THIS_WINDOW lines are new turns in the current sliding window.

# Task
- Segment the current sliding window into atomic, self-contained memory units. Each memory unit should express exactly one independently retrievable claim: one event, state, plan, preference, or other information.
- If the same turn supports multiple independent claims, output multiple memory units with the same start_turn/end_turn when needed. A unit may cover multiple turns only when those turns are needed to make the single claim complete.

## Evidence bound
- Each memory unit MUST include start_turn and end_turn chosen only from source_ids shown in MEMORY_CONTENT.

## Memory-unit fact_text
- fact_text is compact answer evidence. It must be specific enough to answer a later question without reading MEMORY_CONTENT.
- Resolve pronouns and speaker-relative references into the most specific supported entity, role, object, or group. Prefer real names when available; use `the user` or `the assistant` only when no more specific identity is given.

## Triple extraction
- Output one `triple` for the core claim of each memory unit. It must be directly supported by the unit and include subject, predicate, and object.

# Output format
Return valid JSON only. If there is no retrievable factual content with non-overlap evidence, return {"memory_units": []}.
{
  "memory_units": [
    {
      "start_turn": "source id from MEMORY_CONTENT",
      "end_turn": "source id from MEMORY_CONTENT",
      "fact_text": "standalone reusable memory-unit text",
      "triple": {
        "subject": "...",
        "predicate": "...",
        "object": "..."
      }
    }
  ]
}

[User template]
Construct reusable memory units from this fixed sliding-window dialogue segment.

Session id: {session_id}
Turn range: {start_turn}-{end_turn}
Utterance time: {utterance_time}

MEMORY_CONTENT:
{annotated_memory_content}

Output schema:
{
  "memory_units": [
    {
      "start_turn": "...",
      "end_turn": "...",
      "fact_text": "...",
      "triple": {
        "subject": "...",
        "predicate": "...",
        "object": "..."
      }
    }
  ]
}
\end{lstlisting}
\end{tcolorbox}

\begin{tcolorbox}[
  enhanced,
  breakable,
  colback=gray!10,
  colframe=black,
  title={Memory-Unit Retrieval Index Prompt},
  fonttitle=\bfseries\footnotesize,
  width=\columnwidth,
  left=1mm,
  right=1mm,
  top=1mm,
  bottom=1mm,
  boxrule=0.5pt,
  before skip=6pt,
  after skip=6pt
]
\begin{lstlisting}[style=promptstyle,basicstyle=\rmfamily\footnotesize]
[System]
You extract retrieval indexes for memory-unit candidates from one dialogue segment.

Input contains a list of memory units. Each unit has memory_unit_id, utterance_time, fact_text, and evidence_text.

For each memory unit:
1) Extract entities: explicit people, named entities, events, activities, or concrete object/topic keywords needed to retrieve this unit and distinguish it from other memories.
2) Extract locations: concrete places, venues, facilities, cities, countries, or regions where the fact happened. Do not put events or activities in locations unless the text names a physical venue.
3) Extract time: when the described fact happened. Each item is {"start": "...", "end": "...", "granularity": "day|week|month|year"}. Use [] for unknown or vague recency.

Time format:
- For an exact date, set start=end and use the matching granularity.
- For a relative but bounded time, such as "last weekend" or "two weeks before DATE", output a conservative date range rather than a guessed day.
- Distinguish utterance_time and fact time. Resolve relative expressions against the memory unit's own utterance_time, and do not copy utterance_time itself as the fact time.
- For vague times such as "recently", "lately", "a while ago", or "the other day", output [].

Return valid JSON only:
{
  "memory_units": [
    {
      "memory_unit_id": "...",
      "entities": ["..."],
      "locations": ["..."],
      "time": [
        {
          "start": "YYYY-MM-DD|YYYY-MM|YYYY",
          "end": "YYYY-MM-DD|YYYY-MM|YYYY",
          "granularity": "day|week|month|year"
        }
      ]
    }
  ]
}

[User template]
Extract retrieval indexes for these memory units from one dialogue segment.

Session id: {session_id}
Turn range: {start_turn}-{end_turn}
Segment utterance time: {utterance_time}

MEMORY_UNITS:
[
  {
    "memory_unit_id": "...",
    "utterance_time": "...",
    "fact_text": "...",
    "evidence_text": "..."
  }
]
\end{lstlisting}
\end{tcolorbox}

\begin{tcolorbox}[
  enhanced,
  breakable,
  colback=gray!10,
  colframe=black,
  title={Dynamic Memory Trace Consolidation Prompt},
  fonttitle=\bfseries\footnotesize,
  width=\columnwidth,
  left=1mm,
  right=1mm,
  top=1mm,
  bottom=1mm,
  boxrule=0.5pt,
  before skip=6pt,
  after skip=6pt
]
\begin{lstlisting}[style=promptstyle,basicstyle=\rmfamily\footnotesize]
[System]
You are responsible for turning candidate atomic memories into one concise memory trajectory. Several memory units may describe the same subject-relation, and later memories can update, refine, repeat, conflict with, or add to earlier memories. The input items are candidate members of one relation cluster, but some may be irrelevant or only loosely related.

Write one chronological trajectory supported by the memory units:
- keep only items that describe the same evolving subject-relation,
- order the relevant states by memory time,
- include both earlier history and the latest/current supported state.

Use memory order, not world knowledge:
- If time values are `serial-N`, larger N means newer memory.
- If a later memory contradicts an earlier one, describe the later memory as an update.

Return only valid JSON:
{
  "trajectory": "compact chronological trajectory including earlier state(s) and the latest supported state"
}

[User template]
Write one memory trajectory for the following relation cluster.

RELATION:
{relation_text}

MEMORY_UNITS are sorted from oldest to newest by memory time. Write the trajectory from earlier memory to the latest supported memory, including updates and non-conflicting additions.
{members_json}
\end{lstlisting}
\end{tcolorbox}

\begin{tcolorbox}[
  enhanced,
  breakable,
  colback=gray!10,
  colframe=black,
  title={Query Intent Parsing Prompt},
  fonttitle=\bfseries\footnotesize,
  width=\columnwidth,
  left=1mm,
  right=1mm,
  top=1mm,
  bottom=1mm,
  boxrule=0.5pt,
  before skip=6pt,
  after skip=6pt
]
\begin{lstlisting}[style=promptstyle,basicstyle=\rmfamily\footnotesize]
[System]
You parse a question into retrieval constraints for a long-term memory store.

# Background
- The memory store indexes each memory unit by `entities`, `locations`, and `time`.
- Your output is used to select candidate memories before answer generation.
- The extracted values are used for semantic retrieval, so they should be discriminative rather than general.

# Task
- Read the question and extract retrieval constraints on three axes: `entities`, `locations`, and `time`.
- Keep each list concise and deduplicated.
- If an axis has no useful constraint, return an empty list for that axis.

# Entity and Location
- Extract people, named objects, events, activities, roles, organizations, artifacts, and concrete topic words in `entities`.
- Extract physical or named places, such as Paris, Pacific Northwest, school, gym, museum, beach, or library, in `locations`.
- Do not put answer types in `locations`. For example, "countries", "states", "places", or "where" describe what the answer should find.
- Do not put events or activities in `locations`; keep them in `entities` when they constrain evidence.

# Time
- `time.fact` is for when the remembered fact happened, and `time.utterance` is only for when the conversation happened, such as "in our chat on June 3" or "as mentioned last Tuesday".
- Normalize clear dates as date-only values: `YYYY-MM-DD` for a day or `YYYY-MM-DD..YYYY-MM-DD` for an interval.

[User template]
Parse this query into retrieval-oriented intent.

Query:
{query}

Question time:
{question_time}

Use Question time only as the reference for explicit relative time expressions in the query. Do not output it as a time constraint by default.

Output schema:
{
  "entities": ["explicit person, named entity, or topic/object constraint"],
  "locations": ["concrete place or venue constraint stated in the query"],
  "time": {
    "fact": ["YYYY-MM-DD or YYYY-MM-DD..YYYY-MM-DD for when the remembered fact happened"],
    "utterance": ["YYYY-MM-DD or YYYY-MM-DD..YYYY-MM-DD only when the query explicitly constrains when the conversation happened"]
  }
}
\end{lstlisting}
\end{tcolorbox}

\begin{tcolorbox}[
  enhanced,
  breakable,
  colback=gray!10,
  colframe=black,
  title={Final Answer Prompt},
  fonttitle=\bfseries\footnotesize,
  width=\columnwidth,
  left=1mm,
  right=1mm,
  top=1mm,
  bottom=1mm,
  boxrule=0.5pt,
  before skip=6pt,
  after skip=6pt
]
\begin{lstlisting}[style=promptstyle,basicstyle=\rmfamily\footnotesize]
[System]
You answer questions using only the provided memory context.

# Input
- QUESTION is the user's question.
- QUESTION_TIME, when provided, is the time the question was asked. Use it only to resolve relative time expressions in the question.
- MEMORY_CONTEXT includes:
  - Evidence Clusters: retrieved memories grouped by memory-evolution relation. A cluster includes a `trajectory` field describing earlier history and the latest state, together with labeled memory-unit text.
  - Facts: labeled structured subject-predicate-object facts.
- Memory-unit `utterance_time` is when the dialogue happened; if memory-unit times use `serial-N`, larger N means newer evidence.

# Task
- Produce the best supported answer to QUESTION.
- Ground the answer only in MEMORY_CONTEXT.
- When a trajectory's latest state conflicts with older content or older Facts from the same relation, use the trajectory's latest state as the latest memory-supported evidence.

# Output Format
Think step by step internally, then return valid JSON only:
{
  "reasoning_path": "brief chain of reasoning citing labels (MU#/F#) used",
  "answer": "final answer string only"
}

[User template]
Question:
{query}

Question time:
{question_time}

Graph:
{serialized_graph}

Return only valid JSON with keys reasoning_path and answer, in that order.
\end{lstlisting}
\end{tcolorbox}